\documentclass{article}

\usepackage[preprint]{neurips_2025}
\setcitestyle{numbers,square,comma}

\usepackage[utf8]{inputenc}
\usepackage[T1]{fontenc}

\usepackage[colorlinks=true, allcolors=blue]{hyperref}

\usepackage{url}
\usepackage{booktabs}
\usepackage{amsfonts}
\usepackage{nicefrac}
\usepackage{microtype}
\usepackage{xcolor}
\usepackage{subcaption}
\usepackage{wrapfig}
\usepackage{capt-of}

\usepackage[english]{babel}

\usepackage{CJKutf8}

\usepackage{amsmath}
\usepackage[pdftex]{graphicx}

\usepackage{tabularx}
\usepackage{amssymb}
\usepackage{algorithm}
\usepackage{algorithmic}
\usepackage[commandnameprefix=always]{changes}
\usepackage{mdframed}
\usepackage{multirow}
\usepackage{makecell}
\usepackage{appendix}
\usepackage{xparse}
\usepackage{textpos}

\title{From Large to Super-Tiny: End-to-End Optimization for Cost-Efficient LLMs}

\author{
  Jiliang Ni\textsuperscript{*} \quad Jiachen Pu\textsuperscript{*} \quad Zhongyi Yang\textsuperscript{*} \quad Kun Zhou \quad Hui Wang \\
  \textbf{Xiaoliang Xiao \quad Dakui Wang \quad Xin Li \quad Jingfeng Luo \quad Conggang Hu\textsuperscript{\dag}} \\
  Intelligent Connectivity \\
  Alibaba Group \\
  \texttt{conggang.hcg@alibaba-inc.com}
}

\begin{document}

\renewcommand{\sectionautorefname}{Section}
\renewcommand{\subsectionautorefname}{Section}
\renewcommand{\subsubsectionautorefname}{Section}
\renewcommand{\appendixautorefname}{Appendix}

\NewDocumentCommand{\blfootnote}{o m}{
  \begingroup
  \IfValueTF{#1}{
    \IfEqCase{#1}{
      {preprint}{
        \renewcommand\thefootnote{}\footnote{#2}
        \addtocounter{footnote}{-1}
      }
      {final}{
        \renewcommand\thefootnote{}\footnote{#2}
        \addtocounter{footnote}{-1}
      }
    }{}
  }{}
  \endgroup
}

\begin{CJK}{UTF8}{gbsn}

\maketitle

\blfootnote[preprint]{\textsuperscript{*} Equal Contribution }
\blfootnote[preprint]{\textsuperscript{\dag} Corresponding Author}

\begin{abstract}

Large Language Models (LLMs) have significantly advanced artificial intelligence by optimizing traditional Natural Language Processing (NLP) workflows, facilitating their integration into various systems. Many such NLP systems, including ours, directly incorporate LLMs. However, this approach either results in expensive costs or yields suboptimal performance after fine-tuning. In this paper, we introduce a three-stage cost-efficient end-to-end LLM deployment pipeline, comprising prototyping, knowledge transfer, and model compression, to effectively tackle the cost-performance dilemma in LLM-based frameworks. Its high cost-efficiency is manifested not only in simplifying system complexity and producing super-tiny online models with enhanced performance and reduced costs in the results, but also in addressing development cycle constraints, the lack of extensive high-quality data, and limited computational resources during the project development process. In the first stage, we construct an optimal performance prototype system by transforming complex tasks into a function call-based LLM-driven pipeline, which serves as a teacher model to generate high-quality data. In the second stage, we combine techniques like rejection sampling fine-tuning, reinforcement learning, and knowledge distillation to transfer knowledge to 0.5B student models, delivering effective performance at minimal cost. In the final stage, we further compress models to 0.4B via quantization and pruning, achieving ultra-low latency and cost. Extensive experimental results and the framework's modular design suggest cross-domain capabilities and potential applicability in other NLP areas.

\end{abstract}

\section{Introduction}

In recent years, Large Language Models (LLMs) have made significant strides \cite{guo2021selfattentionexternalattentionusing,thoppilan2022lamdalanguagemodelsdialog,akbiketal2018contextual} in enhancing artificial intelligence by addressing shortcomings of traditional pipelines --- including manual annotation, suboptimal performance, and limited generalization and adaptability. Consequently, LLM-based pipelines have stimulated substantial demand for integration into various Natural Language Processing (NLP) systems. We find that numerous such NLP systems, including our own, have directly utilized APIs of closed-source LLMs or fine-tuned open-source LLMs, which can be referred to as the "one-stage" pipeline. Although this pipeline offers implementation simplicity, it suffers from substantial costs \cite{narayanan2021efficientlargescalelanguagemodel} and high inference latency \cite{crossdatasetposeestimation} or suboptimal performance, which limit the broader applicability of LLMs and impact their commercial viability. Within our NLP system, cost-performance is an especially critical metric, a sentiment likely shared by others in the field \cite{gusak2022surveylargescaleneural}. Moreover, our analysis indicates that this pipeline struggles to balance the cost and the performance due to insufficient high-quality data and monotonous optimization techniques.

\begin{figure}
\centering
\includegraphics[width=1.0\linewidth]{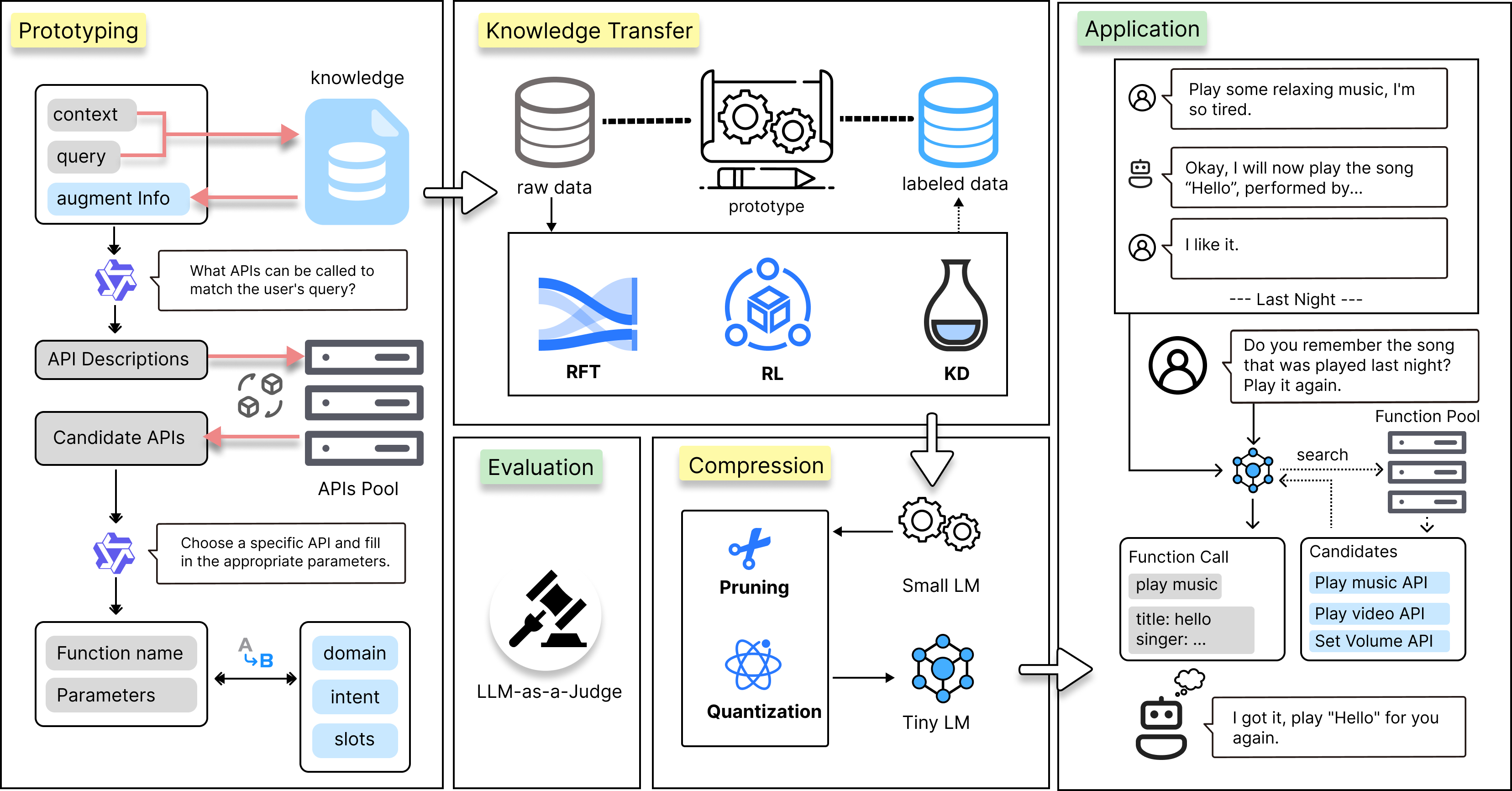}
\begin{subfigure}{0.27\linewidth}
\caption{}\label{image_prototype}
\end{subfigure}
\begin{subfigure}{0.43\linewidth}
\caption{}\label{image_production}
\end{subfigure}
\begin{subfigure}{0.3\linewidth}
\caption{}\label{image_application}
\end{subfigure}
\caption{\label{fig:agent_1}Workflows of prototyping, knowledge transfer and compression.}
\end{figure}

In this paper, we introduce a three-stage cost-efficient end-to-end LLM deployment pipeline, as shown in \autoref{fig:agent_1}, consisting of prototyping, knowledge transfer, and model compression, to address the pervasive "cost-performance" dilemma in the practical deployment of LLM-based frameworks. The high cost-efficiency of this approach is demonstrated by both the outcomes and the process. In terms of outcomes, the approach simplifies system complexity and produces super-tiny models with optimal performance and reduced inference costs and latency. Regarding the process, the approach addresses challenges such as constrained development cycles, the scarcity of substantial high-quality data, and limited computational resources.

In our proposed pipeline, the first stage involves constructing the prototype system, where traditional complex tasks are converted into function call-based LLM-driven workflows \cite{li2025midasmultilevelintentdomain} through prompt engineering, resulting in an optimal performance prototype compared to traditional pipeline. This prototype serves as a teacher model in the second stage, efficiently generating vast quantities of high-quality data. In the second stage, leveraging the optimal prototype from the first stage, we employ techniques including Rejection sampling Fine-Tuning (RFT), Reinforcement Learning (RL) \cite{OpenReasonerZero2025,xie2025logicrlunleashingllmreasoning,tinyzero}, and multi-strategy knowledge distillation (KD) on the Qwen 2.5 series model \cite{qwen2025qwen25technicalreport}. Our hybrid strategy effectively transfers generalization capabilities and the abilities of generating high-quality CoT to a 0.5B student model. The derived student model is considerably smaller than the prototype model while achieving nearly comparable performance at minimal cost. In the third stage, we utilize quantization and pruning techniques to further compress the student model developed in the second phase, resulting in a remarkably compact 0.4B model. Ultimately, this super-tiny model is deployed in our online system. This model achieves hundreds-fold compression and extremely low latency and cost in exchange for a trivial and acceptable loss in performance. The modular design and superior cross-domain performance of the framework suggest its potential applicability in other NLP domains. Our major contributions are as follows:

\begin{itemize}
    \item The first validated three-stage cost-efficient end-to-end LLM pipeline, simplifying system complexity, achieves SOTA results and resolves the "cost-performance" dilemma. It realizes 180 times compression with nearly consistent performance compared to original LLMs and up to an absolute 14\% performance increase over traditional BERT-based systems.
    \item We propose the first hybrid knowledge transfer method by combining RL and KD. It effectively transfers the capabilities of LLMs to super-tiny models and achieves performance comparable and even superior to certain larger models on domain-specific tasks.
    \item The proposed framework offers advantages such as low training, inference, and deployment costs, minimal project development expenses with fast cycles, and a strong potential for cross-domain transferability. It provides inspiration for developing standardized cross-domain solutions in both academia and industry, facilitating the widespread LLMs adoption.
\end{itemize}

\section{Related Works}

\paragraph{LLM for Function Calling \& Evaluation}
Leveraging in-context learning and reasoning capabilities, LLMs can select functions and parameters from provided lists based on user intent \cite{DBLP:conf/nips/SchickDDRLHZCS23}. Fine-tuned 7B models like Gorilla \cite{DBLP:conf/nips/PatilZ0G24} achieve GPT-4-level Function Call accuracy on datasets with thousands of APIs. Building on this, TinyAgent \cite{erdogan2024tinyagentfunctioncallingedge} enables function calling capabilities for edge devices with sub-1.1B parameter models. Query-to-API reformulation via rejection sampling can also optimize API retrieval, reducing retrieval latency while improving accuracy \cite{kachuee2024improvingtoolretrievalleveraging}. Nowadays, LLMs can be used to evaluate model-generated content \cite{DBLP:conf/eamt/KocmiF23,DBLP:conf/nips/ZhengC00WZL0LXZ23}. CriticGPT \cite{mcaleese2024llmcriticshelpcatch} indicates that LLMs can build a cost-effective and high-precision judge, surpassing humans in detecting subtle errors.

\paragraph{Knowledge Distillation}

Knowledge distillation (KD) \cite{DBLP:journals/corr/HintonVD15} transfers knowledge from a teacher model to a smaller student model by simulating  probability distribution. Common approaches include Logits-based\cite{DBLP:conf/iclr/Gu0WH24,dpkd}, Feature-based \cite{yang2023categoriesresponsebasedfeaturebasedrelationbased} and sequence-level (Seq-KD) \cite{DBLP:conf/emnlp/KimR16} distillation. Initially, Logits-based KD uses the forward KL divergence (FKL) \cite{kim2023tokenscaledlogitdistillationternary,sanh2020distilbertdistilledversionbert} as the loss function. Some KD methods prefer Reverse KL divergence (RKL) \cite{gu2024minillmknowledgedistillationlarge,agarwal2024onpolicydistillationlanguagemodels,kim2024promptkddistillingstudentfriendlyknowledge,dpkd}. Recently, Adaptive Kullback-Leiber (AKL) divergence \cite{DBLP:conf/coling/WuTWY0W25} achieves a balance between mimicking distribution and avoiding overfitting.

\paragraph{LLM Pruning \& Quantization}

Efficient LLM deployment primarily relies on pruning\cite{frantar2023sparsegptmassivelanguagemodels,ma2023llmprunerstructuralpruninglarge} and quantization\cite{dettmers2022llmint88bitmatrixmultiplication,frantar2023gptqaccurateposttrainingquantization,xiao2024smoothquantaccurateefficientposttraining}. Structured pruning\cite{li2017pruningfiltersefficientconvnets} employs layer removal\cite{men2024shortgptlayerslargelanguage, muralidharan2024compactlanguagemodelspruning} or dimensionality reduction\cite{ma2023llmprunerstructuralpruninglarge, muralidharan2024compactlanguagemodelspruning,xia2024shearedllamaacceleratinglanguage} using gradient-based\cite{ma2023llmprunerstructuralpruninglarge,xia2024shearedllamaacceleratinglanguage} or activation-aware\cite{muralidharan2024compactlanguagemodelspruning} importance criteria, implemented via learnable masks\cite{xia2024shearedllamaacceleratinglanguage,guo2023compressostructuredpruningcollaborative} or pruning-retraining pipelines\cite{men2024shortgptlayerslargelanguage,ma2023llmprunerstructuralpruninglarge,muralidharan2024compactlanguagemodelspruning}. Weight-only LLM quantization methods\cite{dettmers2022llmint88bitmatrixmultiplication,frantar2023gptqaccurateposttrainingquantization,lin2024awqactivationawareweightquantization} improve memory efficiency while joint weight-activation schemes\cite{xiao2024smoothquantaccurateefficientposttraining,kuzmin2024fp8quantizationpowerexponent} and their combination \cite{lin2024qservew4a8kv4quantizationcodesign} accelerate computation.

\paragraph{Reinforcement Learning}

Recent works \cite{OpenReasonerZero2025,xie2025logicrlunleashingllmreasoning,tinyzero} focus on leveraging Reinforcement Learning (RL) to enhance model reasoning capabilities. Notable reasoning models such as DeepSeek-R1 \cite{deepseekai2025deepseekr1incentivizingreasoningcapability}, QWQ-32B \cite{qwq32b}, and OpenAI o1 \cite{DBLP:journals/corr/abs-2412-16720} have emerged, employing RL algorithms like Proximal Policy Optimization (PPO) \cite{DBLP:journals/corr/SchulmanWDRK17}, Group Relative Policy Optimization (GRPO) \cite{DBLP:journals/corr/abs-2402-03300}, and others \cite{DBLP:conf/acl/AhmadianCGFKPUH24,DBLP:conf/icml/LiXZL00L24,hu2025reinforcesimpleefficientapproach}. For general-purpose LLMs, preference optimization algorithms \cite{DBLP:conf/nips/RafailovSMMEF23,DBLP:conf/icml/EthayarajhXMJK24,DBLP:conf/emnlp/HongLT24,DBLP:conf/nips/0001X024} are widely adopted. Additionally, self-improving pipelines \cite{DBLP:conf/icml/ChenDYJG24,DBLP:journals/corr/abs-2406-01660,DBLP:journals/corr/abs-2408-06195} have been proposed to iteratively refine model performance. We primarily utilize RL to enhance the model's generalization. Compared to directly distilling a reasoning model, RL results in shorter and more precise Chains-of-Thought (CoT).

\section{Methodology}

Recent advancements in LLMs have facilitated significant progress in NLP tasks such as classification and summarization. We harness these capabilities to reconstruct the traditional dialogue system through a cost-efficient end-to-end LLM pipeline, which includes prototyping, knowledge transfer, and model compression, as illustrated in \autoref{image_prototype} and \autoref{image_production}. Initially, we develop the prototype system by leveraging the exceptional performance of contemporary LLMs. Subsequently, we perform knowledge transfer from large models to smaller models. Finally, we achieve additional reduction in model size through further model compression techniques. Ultimately, our objective is to deliver intelligent user experiences at minimal costs, as demonstrated in \autoref{image_application}.

\subsection{Prototyping}\label{sec:prototype}

\paragraph{LLM Prototype}

Traditional dialogue systems usually employ the Intent Classification and Slot Filling (ICSF) framework, often utilizing BERT\cite{chen2019bertjointintentclassification,DBLP:conf/epia/TavaresASSM23}. Given the limitations of a single model in handling complex real-world issues, a Domain Classifier (DC) is often utilized for initial query categorization to facilitate domain-specific ICSF parsing. Although the hierarchical DC-ICSF architecture improves accuracy, the system complexity increases with feature updates. This leads to high inter-module coupling, increasing development and maintenance costs. Furthermore, ambiguous domain boundaries turn the DC into a bottleneck for overall end-to-end accuracy, creating a "feature expansion - operational difficulty" feedback loop that hinders dialogue experience optimization.

Modern LLMs exhibit adequate capability to tackle these NLP problems. Particularly in related Tool Use works, LLMs can learn to call functions without additional training by attaching function descriptions in prompts. Essentially, ICSF can be viewed as a function call task, where the model extracts critical information from user queries as function parameters for specific modules to execute. Following this approach, we design a prototype system based on Qwen2.5-72B-Instruct\cite{qwen2025qwen25technicalreport}. As illustrated in \autoref{image_prototype}, the prototype system accepts user queries and contextual information (including chat history and temporal status information) as input, searches relevant APIs from the API pool, then outputs function names and parameters to call. The traditional system comprises one DC module and multiple ICSF models (based on BERT or regular expressions), each requiring separate maintenance and frequent DC updates. In contrast, the LLM-based system, equipped with function call capability, processes all tasks through a unified model, thereby reducing system complexity and enhancing operational efficiency.

To enhance temporal awareness and mitigate hallucinations, we enrich the input with contextual knowledge via RAG \cite{lewis2021retrievalaugmentedgenerationknowledgeintensivenlp}, retrieving explanations and extensions of entity-related terms from a curated knowledge base. Instead of adding all callable APIs from the API pool into the LLM prompt, we implement an API retrieval module inspired by \cite{kachuee2024improvingtoolretrievalleveraging}. The LLM first generates potential API descriptions, then retrieves top-K APIs via vector similarity (using GTE-Qwen2-1.5B \cite{li2023towards} as the embedding model and FAISS \cite{johnson2019billion} for indexing). Then, the LLM selects optimal functions/parameters from candidate APIs. For backward compatibility, we map functions/parameters to legacy domain-intent-slot structures.

\paragraph{Judge}\label{section:judge}

Traditional ICSF evaluation relies on test sets demanding strict intent-slot alignment, contrasting with real-world variability where utterances permit multiple valid interpretations. For example, "Slightly increase the volume" allows alternative API calls, and "11:00 pm"/"23:00" are semantically equivalent. This inherent ambiguity and complex business logic significantly impede consistent human annotation for accurate ground truth.

To address issues above, we introduce LLM-as-a-Judge to evaluate prototype accuracy. We design a judge based on QwQ-32B\cite{qwq32b}: Given context and a set of Function \& parameters, the judge determines whether this combination semantically satisfies the user query. Following Samuylova's best practices\footnote{https://www.evidentlyai.com/llm-guide/llm-as-a-judge}, we optimize the judge using a 1k-expert-annotated dataset. Subsequently, both the LLM and 9 domain experts evaluate 3k test samples under identical conditions. Results show that judge can provide logically coherent Chain-of-Thought(CoT) rationales and it achieves 93\% consistency with expert judgments. Based on these findings, we conclude that LLM-as-a-Judge is the optimal evaluation approach for this task. We reconstruct the dialogue pipeline from BERT-based DC-ICSF architecture to LLM-based Function Call architecture. Evaluation through LLM-as-a-Judge demonstrates that the new architecture improves accuracy from 81.43\% to 94.74\% compared to the original one.

\subsection{Knowledge Transfer}
\paragraph{Rejection sampling Fine-Tuning}\label{section:sft}

Building on our agent-oriented large model design, we propose Supervised Fine-Tuning (SFT) on smaller models using large model-generated data instead of relying solely on in-context learning (ICL) with large models. This approach offers two key benefits: (1) Smaller models can achieve performance comparable to large models after trained on sufficient high-quality data; (2) It significantly boosts inference and optimization efficiency. For a dataset $D$ containing prompt component $x$ and response component $y$, we define $D(x, y)=\lbrace x_i, y_i\rbrace_{i=1}^{N}$, where $N$ denotes the dataset size.

SFT employs Cross Entropy (CE) loss \cite{HintonSalakhutdinov2006b} as the optimization objective. Building on this, we apply Rejection sampling Fine-Tuning (RFT), utilizing an LLM judge, as introduced in Section \ref{section:judge}, to assess training data generated by the LLM prototype, filter out low-quality data, and construct the final training dataset.
The actual training data is formulated as:

\begin{equation}
D_{filter} = \lbrace x_i, y_i \lvert \ \text{judge\_data}(x_i, y_i) = 1 \rbrace_{i=1}^{N}
\end{equation}

where

\begin{equation}
\text{judge\_data}(x, y)=
\begin{cases}
  1, & \text{if the LLM judge thinks}\ y\ \text{is a good answer for}\ x \\
  0, & \text{otherwise}
\end{cases}
\label{equ:judge_data}
\end{equation}

\paragraph{Knowledge Distillation}\label{sec:distill}

Knowledge Distillation (KD) \cite{DBLP:journals/corr/HintonVD15} offers an alternative to RFT for transferring capabilities from large to small models, typically requiring fewer data samples and exhibiting lower overfitting risks. We adopt the logits-based Adaptive Kullback-Leibler (AKL) distillation \cite{DBLP:conf/coling/WuTWY0W25}:
\begin{equation}
L_{AKL} = \mathbb{E}_{(x,y)\sim D}\lbrack AKL( \pi_\phi(y \lvert x), \pi_\theta(y \lvert x) ) \rbrack
\end{equation}
\begin{equation}
AKL(p, q) = \alpha_{head} KL(p, q) + (1 - \alpha_{head}) KL(q, p)
\end{equation}
\begin{equation}
KL(p, q) = \sum_{t}^{}{p_t}\log\frac{p_t}{q_t}
\end{equation}
where $\pi_\theta$ denotes the trainable student model, $\pi_\phi$ represents the teacher model, and $\alpha_{head}$ is the head ratio in AKL. Setting $\alpha_{head}$ = 1 yields Forward KL Divergence-based distillation \cite{sanh2020distilbertdistilledversionbert}, while $\alpha_{head}$ = 0 corresponds to Reverse KL Divergence-based distillation \cite{dpkd}.

We make optimizations on original methods: (1) Teacher Calibration: Following \cite{sreenivas2024llmpruningdistillationpractice}, we calibrate the teacher model with task-specific data to achieve lower student model loss; (2) Logits Pre-storage: For reducing training time, we pre-store teacher logits to enable multi-student distillation from a single cached logits set; (3) Top-k Pruning: We trim low-probability logits through top-k pruning, mitigating storage pressure and accelerating training speed.

\paragraph{Hybrid Knowledge Transfer}

\begin{wrapfigure}{R}{0.5\textwidth}

    \centering
    \includegraphics[width=\linewidth]{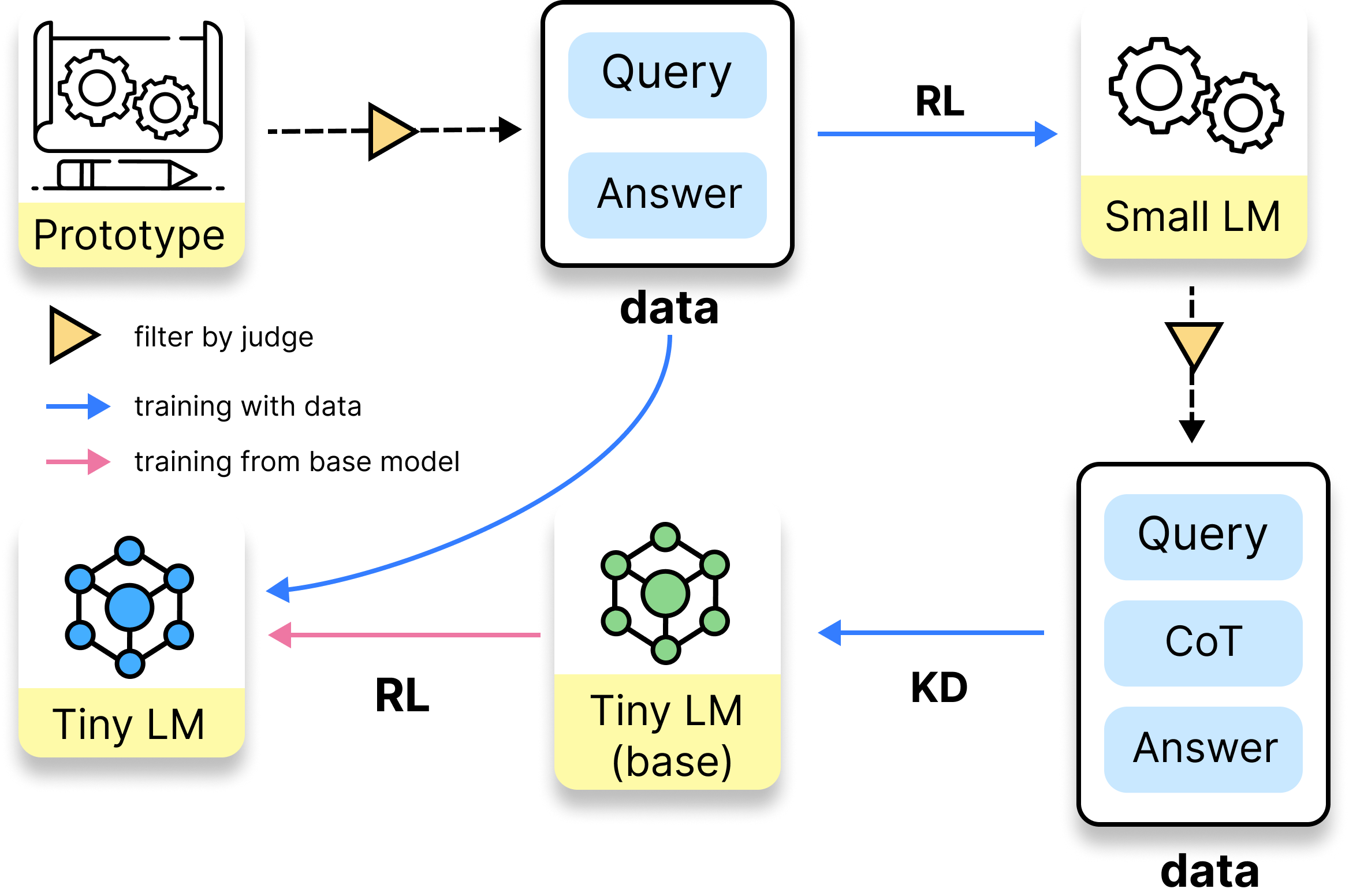}
    \caption{The workflow of the hybrid knowledge transfer strategy.}
    \label{fig:hybrid_kd}
\end{wrapfigure}

Following the standard knowledge transfer strategy, we employ either RFT or KD combined with prompt compression, which removes unnecessary format specifications and Chain-of-Thought (CoT) rationales (see \autoref{sec:prompt_compress}), to create a fast but domain-specific model. However, our experiments reveal limitations of this conventional approach. Utilizing RFT or KD alone can improve inference efficiency but fails to enhance out-of-domain generalization. According to Logic-RL \cite{xie2025logicrlunleashingllmreasoning} with task-specific reward adjustment (see \autoref{sec:rl_details}), we employ Reinforcement Learning (RL) to improve generalization. Additional experiments indicate that employing RL alone generally leads to varying effectiveness across different model sizes: RL training yields exceptional results on small models (7B), surpassing both the LLM prototype and RFT-trained models. On the other hand, when applied to super-tiny models (0.5B), RL becomes ineffective as these models tend to generate shorter or meaningless CoT responses, leading to degraded output diversity.

These findings motivate us to propose a hybrid knowledge transfer strategy, as illustrated in \autoref{fig:hybrid_kd}. This hybrid strategy first employs RL to train models that are as small as possible while ensuring robust generalization and the ability to generate high-quality CoT, exemplified by the 7B RL-trained model in our experiments. Subsequently, KD is used to transfer these generalization capabilities and CoT generation abilities to super-tiny models. With CoT generation abilities, super-tiny models can be trained through a second round of RL to further enhance generalization. Experimental results demonstrate that super-tiny models exhibit impressive improvement in both in-domain and out-of-domain performance after the second RL training. Detailed performance comparison of different transfer strategies shows in \autoref{sec:full_result}.

\subsection{Model Compression}
\paragraph{Pruning}

To further compress model, we design two phases of pruning following NVIDIA \cite{sreenivas2024llmpruningdistillationpractice}, Llama3.2 \cite{meta2024llama32}, and Apple \cite{gunter2024appleintelligencefoundationlanguage}: (1) Offline One-Shot Struct Pruning: Following Minitron\cite{ma2023llmprunerstructuralpruninglarge}, we explore two pruning strategies. For depth pruning, we compute layer importance scores and remove the $N$ least important layers based on the target pruning ratio. For width pruning, we conduct a model architecture search using PPL-based criteria and prune channels according to group-wise importance rankings, with full implementation details in \autoref{sec:prune_detail}; (2) Recovery training: The first step is General Instruction Fine-Tuning via Knowledge Distillation. Adopting Minitron's approach, we leverage logits from a larger general instruction teacher model for recovery training. Due to unavailability of the teacher's original training data, we first perform model correction on our selected IFT dataset to stabilize distillation, followed by top-k logits distillation(\autoref{sec:distill}). The second step is Domain-Specific Task Alignment. The pruned compact LM undergoes further aligning through knowledge distillation, producing a final domain-optimized LLM.

\paragraph{Quantization}

Quantization typically reduces GPU memory consumption and enhances inference performance. In this work, we apply post-training quantization to smaller models obtained through distillation/pruning. Specifically, we implement two quantization approaches: 4-bit weight-only GPTQ algorithm \cite{frantar2023gptqaccurateposttrainingquantization} and full 8-bit weight-activation FP8 quantization \cite{kuzmin2024fp8quantizationpowerexponent}. The 4-bit weight-only quantization \cite{frantar2023gptqaccurateposttrainingquantization} compresses linear weights from FP16 to per-group INT4 via integer quantization, which are dequantized to FP16 during computation. This approach achieves notable inference acceleration in memory-bound scenarios with large model parameters and low GPU memory bandwidth. The FP8 quantization \cite{kuzmin2024fp8quantizationpowerexponent} quantizes both weights and activations to 8-bit floating-point format. Compared to 8-bit integer quantization \cite{kuzmin2024fp8quantizationpowerexponent}, FP8 shows a more noticeable improvement in accuracy, and by utilizing NVIDIA's FP8 tensor cores, FP8 can achieve effective inference acceleration \cite{tensorrt-llm}.

\section{Experiments}
\subsection{Experimental Setup}

\paragraph{Datasets}

In our experiments, we utilize a large-scale dataset built from years of team experience and practice accumulated in the field. We reconstruct the intent classification and semantic slot framework within the real-world system into an API document specification, creating a function pool comprising 74 functional modules with 225 arguments. We sample 200k data points from a single business scenario as the training set and synthesize labels based on the LLM prototype outlined in \autoref{sec:prototype}. A combination of stratified and random sampling yields 11,922 samples from this scenario's data as the in-domain test set. Additionally, we sample 7,867 data points using the same methodology from another business scenario, without some callable functions in the training set, as the out-of-domain test set.

\paragraph{Metrics}

In accordance with the LLM-based judge outlined in \autoref{section:judge}, we establish our evaluation metrics. Consider a dataset of size $N$, denoted as $D(x,y) = \lbrace x_i, y_i\rbrace_{i=1}^{N}$, where $x$ signifies the prompt, including elements such as user queries, function lists, and dialogue histories, while $y$ represents the response containing the function and its parameters. The function judge\_data, as rigorously defined in \autoref{equ:judge_data}, assesses whether the function and its parameter list can fulfill the specified user query. We define the Achievable Rate (AR) as follows:

\begin{equation}\label{equ:achievable_rate}
\text{Achievable Rate (AR)} = \frac{1}{N} \sum_{i=1}^{N} \text{judge\_data}(x_i, y_i)
\end{equation}

\paragraph{Training}

During the training phases, base models are all from small instruct models (0.5B-7B) of the Qwen2.5 model family \cite{qwen2025qwen25technicalreport}. We conduct RFT on them and RL on 7B models. We perform pruning on 0.5B models and obtain Qwen2.5-0.4B-Width and Qwen2.5-0.4B-Depth, both under a 20\% pruning ratio. Then, we transfer knowledge from 7B RFT-trained models to pruned and 0.5B models via KD. We also apply KD from 7B RL-trained models to 0.5B models, followed by another RL phase (denoted as KD\textsuperscript{*}+RL). Finally, we use model quantization via W8A16/W4A16/W8A8 strategies on the KD-trained 0.5B model, detailed in \autoref{sec:train_details}. The training process is cost-efficient, utilizing only 8 NVIDIA H20 GPUs. Computing resources are listed in \autoref{sec:compute_resources}.

\paragraph{Evaluation settings}

We do evaluation by comparing our models with the BERT-based system consisting of two models based on BERT-large (0.33B)\cite{DBLP:conf/naacl/DevlinCLT19}. To evaluate the effectiveness of our method, we invite expert annotators alongside the LLM-as-a-Judge-based metric AR (see \autoref{equ:achievable_rate}) for performance assessment. We also implement a retrieval module to select top 10 functions from the function pool, aiming to accelerate both training and inference phases, as detailed in \autoref{sec:retriever_setup}. To further verify the effectiveness of our proposed approach, we train models on the xlam-function-calling-60k dataset generated by APIGen\cite{DBLP:conf/nips/LiuHZZLKTYLFNYS24} and conduct validation using the Berkeley Function-Calling Leaderboard (BFCL)\cite{berkeley-function-calling-leaderboard}. Details can be found in \autoref{sec:exp_on_pub_datasets}. In real-world scenarios, we deploy models on NVIDIA L20 GPU and perform effectiveness evaluation via TensorRT-LLM\cite{tensorrt-llm}, with specific configuration details provided in \autoref{sec:infer_details}. During the comparison between the trained models and the BERT-based system, we adopt queries per second (QPS) as the primary metric for system throughput, response time (RT) for end-to-end latency, and concurrent clients (CC) to quantify high-concurrency performance.

\subsection{Experimental Results}

\noindent
\begin{minipage}{0.58\linewidth}
\centering
\captionof{table}{\label{tab:main_result}Main results of the models trained on Qwen2.5 (See \autoref{sec:FLOPs} for FLOPs analysis).}
\vspace{\abovecaptionskip}
\begin{tabular}{l|l|r}
\toprule
Model / System & AR & FLOPs (G) \\
\midrule

BERT-based system & 81.43\% & 38.65 \\
Qwen2.5-72B (ICL) & 94.74\% & 4907.50 \\
Qwen2.5-7B (RFT) & \underline{95.29}\% & 434.04 \\
Qwen2.5-3B (RFT) & \textbf{95.30}\% & 218.94 \\
Qwen2.5-1.5B (RFT) & 95.16\% & 111.91 \\
Qwen2.5-0.5B (RFT) & 94.66\% & 43.31 \\
Qwen2.5-0.5B (KD) & 94.85\% & 43.31 \\
Qwen2.5-0.5B-W8A16 (KD) & 94.76\% & - \\
Qwen2.5-0.5B-W4A16 (KD) & 94.19\% & - \\
Qwen2.5-0.5B-W8A8 (KD) & 94.83\% & - \\
Qwen2.5-0.4B-Width (KD) & 94.17\% & 40.83 \\
Qwen2.5-0.4B-Depth (KD) & 94.20\% & 33.10 \\
\bottomrule
\end{tabular}

\end{minipage}
\hfill
\begin{minipage}{0.38\linewidth}
\centering
\includegraphics[width=\linewidth]{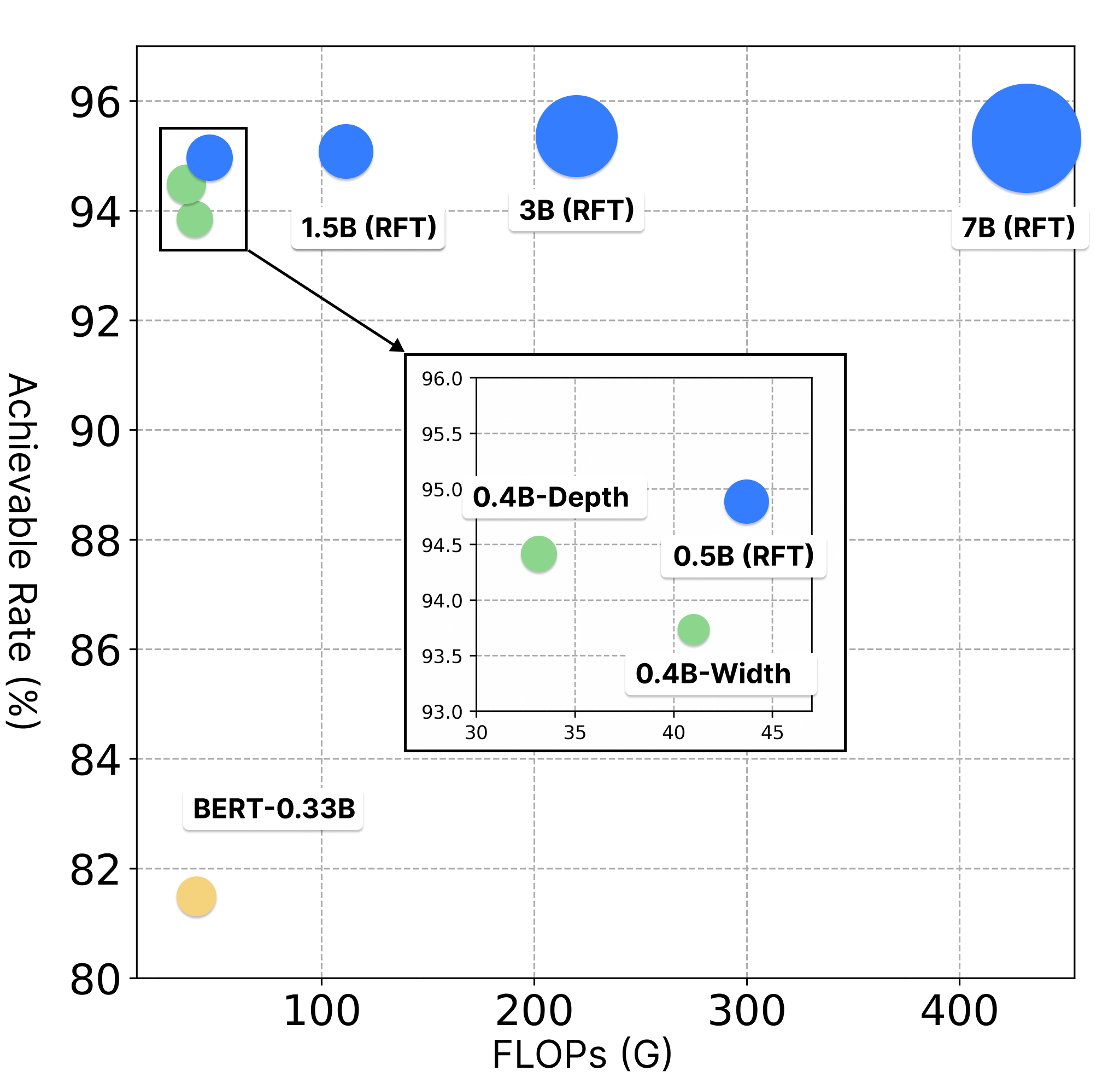}
\captionof{figure}{\label{fig:ar_flops}Cost and performance results. FLOPs reduced by 10x with nearly lossless AR.}
\end{minipage}

\noindent
\begin{minipage}{0.52\linewidth}
\centering
\captionof{table}{\label{tab:sft_vs_rl}In-domain and out-of-domain performances of different knowledge transfer strategies.}
\vspace{\abovecaptionskip}
\begin{tabular}{l|r|r}
\toprule
Model & \multicolumn{1}{p{1cm}}{\centering AR \\ (ID)} & \multicolumn{1}{|p{1cm}}{\centering AR \\ (OOD)} \\
\midrule
Qwen2.5-72B (ICL) & 94.74\% & \underline{94.63}\% \\
Qwen2.5-7B (RFT) & 95.29\% & 90.86\% \\
Qwen2.5-0.5B (RFT) & 94.66\% & 90.62\% \\
Qwen2.5-7B (RL) & \underline{95.36}\% & \textbf{94.80}\% \\
Qwen2.5-0.5B (KD\textsuperscript{*}+RL) & \textbf{95.99}\% & 92.34\% \\
\bottomrule
\end{tabular}
\end{minipage}
\hfill
\begin{minipage}{0.46\linewidth}
\vspace{.8em}
\centering
\includegraphics[width=1.0\linewidth]{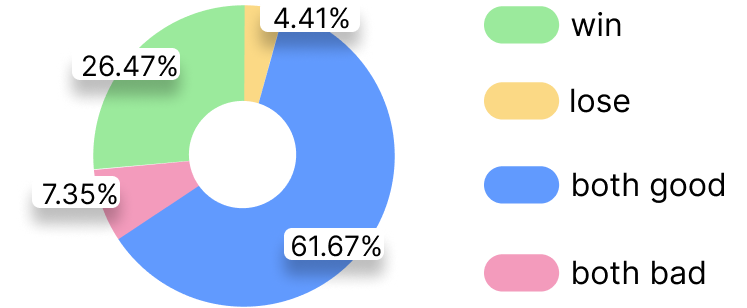}
\captionof{figure}{Comparison of annotation results between the LLM prototype and the BERT-based system determined by professional annotators.}
\label{fig:compare_metrics}
\end{minipage}

\paragraph{LLM Prototype Preferred by Both Human and LLM-as-a-Judge}
As illustrated in \autoref{fig:compare_metrics}, annotations provided by professional annotators reveal that the LLM prototype (72B ICL) achieves a superior performance over the BERT-based system, with a win+tie rate of 95.59\%. Through the evaluation via LLM-as-a-Judge, we find that the LLM Prototype surpasses the BERT-based system by 13.31\%, as shown in \autoref{tab:main_result}. This conclusion is nearly consistent with that presented by professional annotators, demonstrating the effectiveness of the LLM judge.

\paragraph{RFT and KD as Effective Knowledge Transfer Methods}

As demonstrated in \autoref{tab:main_result}, models trained with RFT generally exhibit a higher AR than the LLM prototype. Specifically, both the 3B and 7B models display negligible differences in performance and both achieves a 0.5\% improvement in AR compared to the LLM prototype. The 0.5B model's AR is 0.5\% lower than the 1.5B model, while KD emerges as a valuable complement to RFT, enhancing the performance of the 0.5B model by an additional 0.19\%. In terms of inference performance, the 0.5B model demonstrates only a slight 0.4\% difference in AR compared to the 7B model, while achieving a significant 13.57x improvement in QPS and a substantial 91.3\% reduction in latency, as illustrated in \autoref{tab:compress_infer_v3}.

\paragraph{RL's Superior Generalization Ability}
According to \autoref{tab:sft_vs_rl}, while RFT performs well on in-domain data, it demonstrates worse results on out-of-domain data with a 3.61\% AR decrease compared to the LLM prototype. Conversely, 7B RL-trained models achieve superior performance compared to other models on both in-domain and out-of-domain datasets, indicating better generalization of RL training methodology. Notably, even with models as small as 0.5B, employing a hybrid training methodology with KD+RL can yield superior in-domain results and an improvement of 1.7\% in out-of-domain datasets compared to the RFT method. For more detailed exploration of the RL results, please refer to \autoref{sec:full_result}.

\paragraph{Tiny LLMs as a Cost-Efficient Upgrade of BERT-based System}

\begin{figure}
\centering
\includegraphics[width=0.99\linewidth]{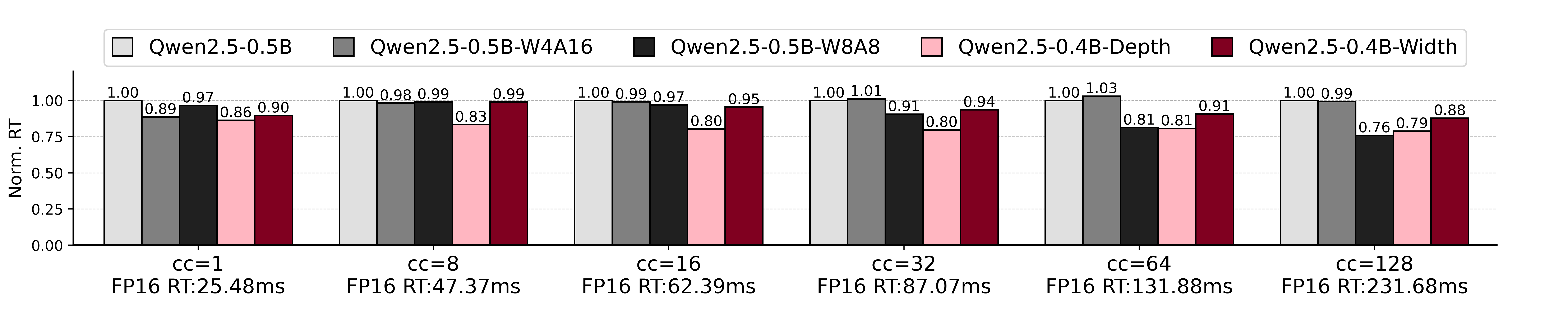}
\caption{\label{fig:rt_ratio}Comparison of RT metrics using different quantization and pruning strategies.}
\end{figure}

As shown in \autoref{tab:main_result}, \autoref{tab:compress_infer_v3} and \autoref{fig:rt_ratio}, quantization minimally impacts accuracy, with less than a 0.1\% AR loss, while significantly enhancing inference performance, achieving up to 1.29x higher QPS and 0.76x lower RT in the optimal scenario. Pruning incurs a slightly higher 0.2\% accuracy loss but consistently enhances performance by 20\%. For end-to-end inference, with a CC of 64, the BERT-based system reaches 1110.88 QPS and 55.84 ms RT. Although the cost is roughly twice that of the BERT-based system, the AR improves by up to absolute 14\%.

\begin{table}[]
\caption{Comparison of QPS metrics using different quantization and pruning strategies.}
\centering
\label{tab:compress_infer_v3}
\begin{tabular}{lcccccc}
\toprule
\multicolumn{1}{c}{\multirow{2}{*}{Model}} & \multicolumn{6}{c}{QPS}                                                                                  \\ \cline{2-7}
\multicolumn{1}{c}{}                       & cc=1           & cc=8            & cc=16           & cc=32           & cc=64           & cc=128          \\ \midrule
Qwen2.5-0.5B                               & 39.11          & 166.71          & 248.62          & 348.87          & 459.12          & 517.13          \\ \midrule
Qwen2.5-0.5B-W8A16                         & 44.11          & 169.86          & 251.21          & 346.17          & 445.84          & 521.41          \\
Qwen2.5-0.5B-W4A16                         & \textbf{49.83} & \textbf{175.29}          & 256.05          & 338.16          & 453.46          & 519.22          \\
Qwen2.5-0.5B-W8A8                          & 40.40          & 158.15 & \textbf{256.32} & \textbf{382.13} & \textbf{558.53} & \textbf{668.35} \\ \midrule
Qwen2.5-0.4B-Width                         & 43.65          & 167.65          & 260.58          & 372.59          & 502.82          & 581.23          \\
Qwen2.5-0.4B-Depth                         & \textbf{45.22} & \textbf{198.82} & \textbf{309.62} & \textbf{438.58} & \textbf{566.87} & \textbf{650.26} \\ \midrule
Qwen2.5-7B                                 & 4.43              & 14.77               & 17.71               & 19.04               &  19.79              & 20.06               \\ \bottomrule
\end{tabular}
\end{table}

\section{Analysis}

\paragraph{Performance-Efficiency Trade-offs in Model Scaling}

Our experimental analysis highlights a compelling trade-off between model size and computational efficiency. Although larger models typically show superior performance, the expected disparity between large and small ones is not as pronounced as anticipated. As shown in \autoref{fig:ar_flops}, the 0.4B model achieves performance almost on par with the 7B model, efficiently requiring 11.8\% FLOPs and reaching more than 17.4x QPS improvement and 92.9\% RT reduction, with the 7B model's RT being 882.552 ms under 16 CC. Although this model has a slight gap in cost compared to traditional BERT models, the improved accuracy and scalability stemming from LLM base makes it a favorable choice for various applications.

\paragraph{Characteristics of Training Paradigms}

The results presented in \autoref{tab:sft_vs_rl} underscore the distinct characteristics of different training methodologies. Fine-tuning primarily focuses on memorizing data, making it susceptible to overfitting to the training set. In contrast, RL presents greater robustness by not merely fitting individual data points. We posit that RL's superior generalization capabilities stem from its ability to learn from negative samples, which are generated by the model being trained itself. RL is highly effective for dealing with well-structured problems, while it has unsatisfactory convergence facing ambiguous data or a lack of domain knowledge in real-world tasks. Additionally, applying RL directly to tiny models often proves to be ineffective, likely due to such models' limited capacity to utilize Chain-of-Thought (CoT) rationales. In such cases, KD can compensate for this limitation by training the tiny models on a large amount of CoT data generated by a RL-trained teacher model. Compared to fine-tuning, KD can better extract CoT rationales from the probability distribution. Consequently, the hybrid methodology of KD and RL can achieve remarkably enhanced performance and represents a promising direction for future development.

\paragraph{Comparison over Model Compression Strategies}
Our analysis identifies depth pruning as the optimal approach for domain-specific models. Depth pruning achieves 20\% inference acceleration with lossless AR at same compression ratios (\autoref{tab:main_result}), stemming from the nature of depth pruning, removing entire transformer layers and their associated attention computations. Although width pruning exhibits slightly better effects in general instruction tasks after training (\autoref{tab:pruned_model_evalation}), depth pruning presents optimal efficiency facing long prompts, with width pruning focusing on matrix optimizations (\autoref{sec:FLOPs}). Regarding quantization, the W4A16 strategy achieves impressive performance in single-thread scenarios, while FP8 outperforms with lossless accuracy under high concurrency conditions as the superior paradigm for throughput deployments.

\section{Conclusion}

Based on our comprehensive study, several key insights emerge regarding the development and deployment of effective LLMs in practical scenarios.

\paragraph{A Scalable Practical Alternative for Cost-Efficient LLMs}
Our proposed pipeline presents a systematic framework designed to achieve high performance while maintaining efficient costs in both development and deployment. Our end-to-end optimized tiny models also outperform their larger counterparts, offering resource efficiency without compromising accuracy. This provides an alternative pathway for applying LLMs in practical tasks under low-cost scenarios. We believe this approach can be effectively adapted to other domains, thereby potentially advancing research and facilitating the widespread deployment of LLMs.

\paragraph{LLM Capability Preservation via a Hybrid Strategy}
Our hybrid strategy demonstrates strong effectiveness and practicality in preserving LLMs' capabilities during knowledge transfer. It enhances generalization and high-quality CoT generation abilities via RL, safely retains these capabilities in smaller models via KD, and further improves the in-domain and out-of-domain performance of super-tiny models via another RL. This methodology judiciously and scientifically avoids irreversible loss of capabilities that might occur when directly deriving super-tiny models via RL. It progressively retains LLMs' capabilities, transferring them step by step to super-tiny models.

\paragraph{Balanced Focus Beyond AGI}
While AGI remains a visionary long-term aim for LLMs, progress has reached a plateau, prompting increased attention towards real-world applicability. Our work demonstrates that integrating LLM-driven techniques into traditional tasks results in immediate, measurable improvements. It advocates for a balanced research agenda that equally emphasizes foundational advancements and domain-specific applications.

\section{Limitation \& Future Works}

Our pipeline has presented optimal cost-performance results, and our super-tiny models have exhibited remarkable capabilities, featuring low latency, cost efficiency, cloud independence, and robust privacy. Further compression may offer limited marginal benefits. Thus, we are more focused on enhancing such models' capabilities and expand their broader applications.

\paragraph{Deeper Study on LLM Capability Transfer}
Based on our experiments, the hybrid strategy effectively enhances super-tiny models' generalization while almost entirely retaining LLMs' capabilities in domain-specific tasks. In future research, we will conduct a deeper investigation and theoretical analysis of this potentially powerful methodology, building upon prior work to gain a better understanding of this hybrid strategy in transferring LLMs' capabilities to super-tiny models.

\paragraph{Cross-domain Optimal Compression Exploration}
It is evident that compressing models to 0.5B while retaining their capabilities across all domains is not feasible. The limitation boundary remains unclear. So, further research and exploration are required to better understand and define these limits.

\newpage
\appendix
\appendixpage

\section{Details for Reinforcement Learning}\label{sec:rl_details}

 Our reward design consists of two components, Format Reward($S_{format}$) and Answer Reward ($S_{answer}$):
\begin{itemize}
    \item For Format Reward, we adopt the logic of Logic-RL\cite{xie2025logicrlunleashingllmreasoning}, which validates the output format against the specified template
    \verb|<think>...</think><answer>...</answer>| via regular expressions. We add additional designs to ensure the validity of the usage of the function and its parameters. If a function not present in the predefined input function pool is selected, it is deemed invalid.
\end{itemize}

\begin{equation}
S_{format} =
\begin{cases}
  1, & \text{if the format is correct, and the function call is valid} \\
  0, & \text{otherwise}
\end{cases}
\end{equation}

\begin{figure}[!h]
\centering
\includegraphics[width=0.8\linewidth]{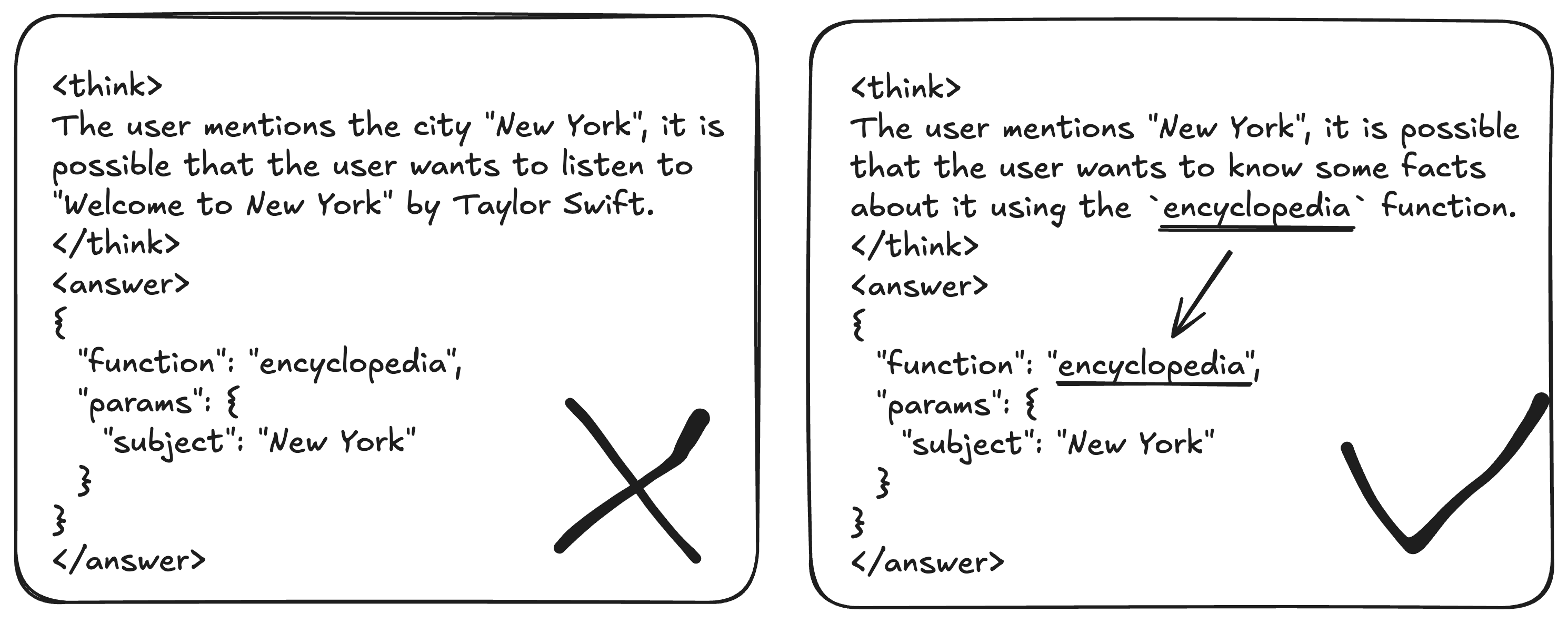}

\caption{\label{fig:rl_case}An example of optimization of the original Answer Reward design by requiring explicit repetition of the answer within the CoT framework.}
\end{figure}

\begin{itemize}
    \item For Answer Reward, in our specific scenario, we observe that the model tends to directly guess answer during training, resulting in no association between the Chain-of-Thought (CoT) and the final results. Requiring explicit repetition of the answer within the CoT framework effectively resolves this issue. As illustrated in \autoref{fig:rl_case}, we optimize its original design in order to enhance the consistency between the Chain-of-Thought (CoT) reasoning and the results. Our requirements are as follows: (1) The CoT must explicitly reference the target function at least once; (2) The final function mentioned in the CoT must correspond to the function in the answer. Failure to comply with these conditions will be regarded as a parsing error. Additionally, we incorporate fine-grained rewards, which provide better guidance when the answer is partially correct, enabling the model to achieve improved performance.
\end{itemize}

\begin{equation}
S_{answer} =
\begin{cases}
  -2 & \text{if the answer is missing, or is unable to be parsed,} \\
  & \text{or is not defined in the function list,} \\
  & \text{or the thought doesn't end with the function in the answer} \\
  -1.25 & \text{if the names of at least one parameter and function are correct,} \\
  -1.2 & \text{if the name of function, as well as the name } \\
  & \text{and value of at least one parameter are correct} \\
  2, & \text{if the answer is correct,} \\
  -1.5, & \text{otherwise}
\end{cases}
\end{equation}

\section{Prompt Compression}\label{sec:prompt_compress}
\begin{figure}[!h]
\centering
\includegraphics[width=1.0\linewidth]{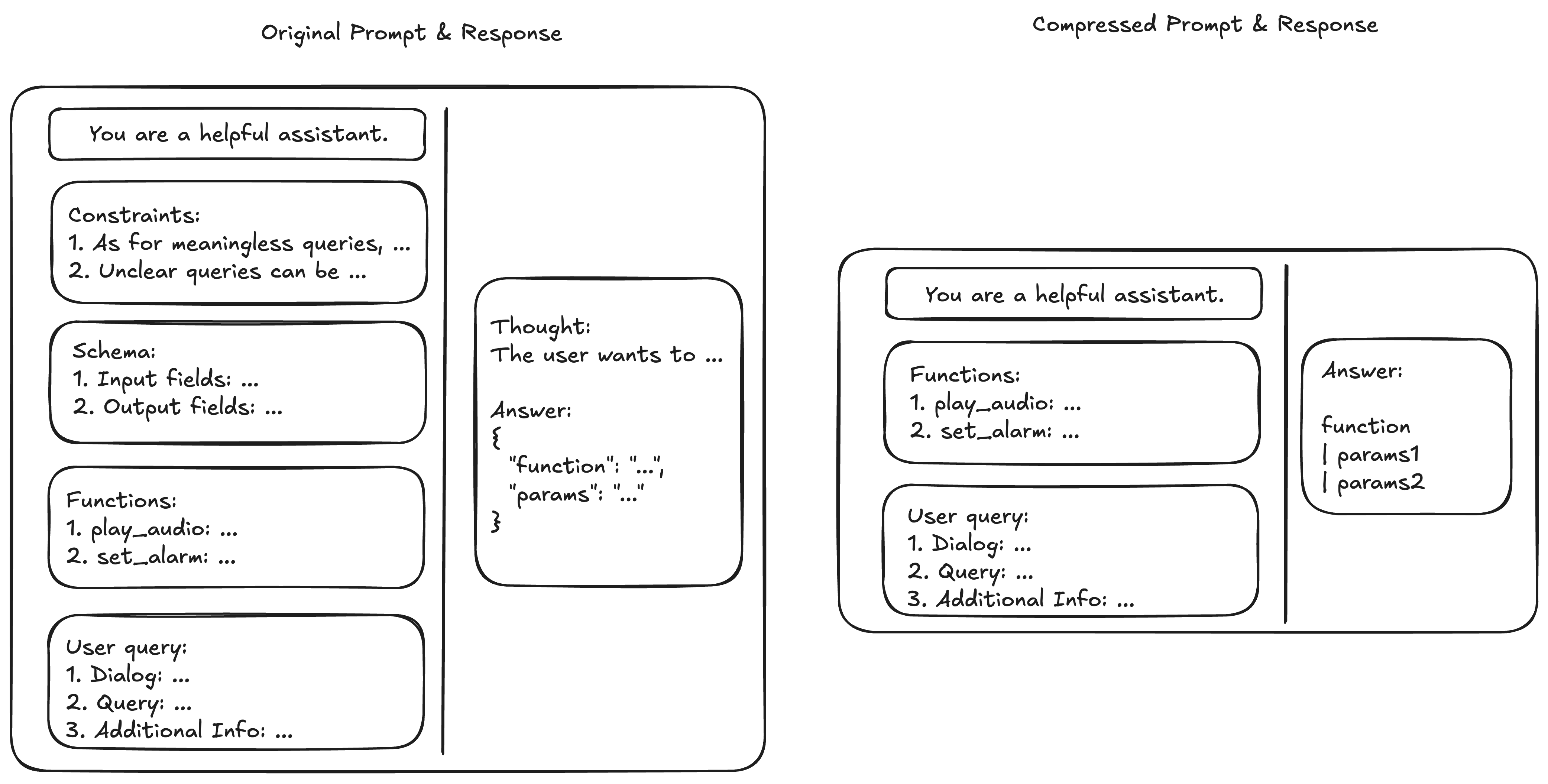}
\caption{\label{fig:prompt_compression}Original vs Compressed Prompt \& Response.}
\end{figure}

To reduce inference costs, we remove unnecessary content in prompts, such as guidelines, format specifications, and few-shot examples used during large model invocation, while retaining essential elements: user queries, context, query-relevant knowledge, function lists, and brief function descriptions. For outputs, we remove Chain-of-Thought (CoT) components and directly use predefined JSON-formatted responses, as illustrated in \autoref{fig:prompt_compression}.

\section{Retriever Setup}\label{sec:retriever_setup}

Initially, we select GTE-large \cite{li2023towards}, which is trained on the BERT-large \cite{DBLP:conf/naacl/DevlinCLT19}, as our embedding model. Using this model, we generate embeddings of API descriptions produced by the LLMs. Then, we compare these embeddings with those of the API description from the document using cosine similarity. However, this model's recall accuracy is not high. Subsequently, we switch to GTE-Qwen2-1.5B-instruct \cite{li2023towards}, a model trained on Qwen2-1.5B \cite{yang2024qwen2technicalreport}, which offers the advantage of fine-tuning for matching tasks through prompt modification. Next, utilizing this model with a customized prompt in \autoref{sec:prompt}, we perform cosine similarity matching between the model-generated API descriptions and the JSON text of the entire API documentation, achieving a higher accuracy. During training, we optimize the API description model using rejection sampling, with implementation details provided in \autoref{sec:retriever_optimization}. Finally, to achieve optimal inference acceleration, we train a classifier based on BERT-large \cite{DBLP:conf/naacl/DevlinCLT19} for top-1 recall using rejection sampling. The classifier takes Dialog, User Query, and necessary context as input and the function category as output. The functions corresponding to the top-k output logits are then considered as candidate functions for the subsequent model's function call. This approach significantly accelerates inference due to the efficiency of BERT classifiers compared to autoregressive GPT models, while also enhancing function recall capabilities. However, it introduces scalability challenges since the recall model needs to be retrained with each update to the function pool. As illustrated in \autoref{table:retrieval_comparison}, this series of optimizations lead to a substantially improvement in API recall rates.

\begin{table}[!h]
\centering
\centering

\caption{Comparison of retrieval performance of different models: GTE-large model, GTE-Qwen2-1.5B-instruct model, optimized API description model + GTE-Qwen2-1.5B-instruct model. Extensibility implies that the API pool can be dynamically modified. }
\label{table:retrieval_comparison}
\begin{tabular}{c|c|c|c|c}
\toprule
\textbf{Top-N} & \textbf{GTE-large} & \textbf{GTE-Qwen2} & \multicolumn{1}{p{2.5cm}|}{\centering \textbf{Opt-API+GTE-Qwen2}} & \textbf{RFT-BERT} \\
\midrule
5 & 60.19\% & 88.27\% & 93.39\% & 99.61\% \\
7 & 74.06\% & 94.67\% & 98.09\% & 99.92\% \\
10 & 90.11\% & 98.22\% & 99.55\% & 99.92\% \\
15 & 98.95\% & 99.80\% & 100\% & 100\% \\
\midrule
\textbf{Extensibility} & √ & √ & √ & × \\
\bottomrule

\end{tabular}
\hfill

\end{table}

\subsection{Prompt for retrieval}\label{sec:prompt}
\begin{mdframed}
Instruction: Given a API description, retrieve the most similar API from the function pool.

Query:
\end{mdframed}

\subsection{Retriever Optimization}\label{sec:retriever_optimization}
\begin{algorithm}[H]
\caption{Simplified API Description Validation Process}
\label{alg:simplified_api_validation}
\begin{algorithmic}[1]
\REQUIRE Query $q$, LLM $\mathcal{M}$, API Pool $\mathcal{R}$, Retrieval Module $\mathcal{S}$, Function Call Module $\mathcal{F}$, Judge $\mathcal{J}$
\ENSURE High-quality training dataset $\mathcal{D}$
\STATE Generate multiple API descriptions $Des = \{\mathcal{M}(q)\}$ in parallel.
\FOR{each description $d \in Des$}
    \STATE Retrieve candidate APIs $C = \mathcal{S}(d, \mathcal{R})$.
    \STATE Select API $a \in C$ and generate parameters $p$.
    \STATE Execute $a$ with $p$, get result $r$.
    \IF{$\mathcal{J}(r) = \text{Success}$}
        \STATE Check if $a$ is in the top 5 candidates of $C$.
        \IF{Yes}
            \STATE Add $(q, d, a, p, r)$ to $\mathcal{D}$.
        \ENDIF
    \ENDIF
\ENDFOR
\end{algorithmic}
\end{algorithm}

\section{Training Details}\label{sec:train_details}

\begin{table}
\parbox{.45\linewidth}{
    \centering
    \caption{Training parameters of different knowledge transfer strategies.}
    \label{tab:experiment_parameters_phases}
    \begin{tabular}{|l|l|}
        \toprule
        \multicolumn{2}{|c|}{\textbf{Supervised Fine-Tuning (SFT)}} \\
        \midrule
        Learning Rate       & $2 \times 10^{-5}$ \\
        Batch Size          & $128$ \\
        Number of Epochs    & $1$ \\
        Weight Decay        & $1 \times 10^{-4}$ \\
        \bottomrule
        \toprule
        \multicolumn{2}{|c|}{\textbf{Reinforcement Learning (RL)}} \\
        \midrule
        Learning Rate       & $3 \times 10^{-7}$ \\
        Batch Size          & $128$ \\
        Number of Episodes  & $1400$ \\
        Number of Rollouts  & $8$ \\
        KL Coefficient      & $0.001$ \\
        \bottomrule
        \toprule
        \multicolumn{2}{|c|}{\textbf{Knowledge Distillation (KD)}} \\
        \midrule
        Learning Rate       & $2 \times 10^{-5}$ \\
        Batch Size          & $128$ \\
        Number of Epochs    & $1$ \\
        Logits Top-K        & $100$ \\
        \bottomrule
    \end{tabular}
}
\parbox{.45\linewidth}{
\centering

\caption{\label{tab:kd_result}Ablation of KD on Qwen2.5-0.5B models.}
\begin{tabular}{l|l}
\toprule
Model & AR \\
\midrule
Qwen2.5-0.5B (RFT) & 94.66\% \\
Qwen2.5-0.5B (FKL) & 94.65\% \\
Qwen2.5-0.5B (AKL[$\mu=0.5$]) & 94.48\% \\
Qwen2.5-0.5B (AKL[$\mu=0.7$]) & \underline{94.68}\% \\
Qwen2.5-0.5B (AKL[$\mu=0.9$]) & \textbf{94.85}\% \\
\bottomrule
\end{tabular}
}
\end{table}

The main training parameters are detailed in \autoref{tab:experiment_parameters_phases}. During the training phase, we employ the AdamW optimizer \cite{DBLP:conf/iclr/LoshchilovH19} with the cosine learning rate schedule, setting the model's maximum sequence length to $5120$. In the RL phase, we utilize the baseline variant of Reinforce++ \cite{hu2025reinforcesimpleefficientapproach} as the reinforcement learning algorithm, offering advantages in efficiency, conciseness, and training stability. Following \cite{Liu2025KL}, the K2 estimator is used instead of the original K1 in Reinforce++. In the AKL phase, the threshold parameter $\mu$ is reduced from 1 to 0.5. When training with data containing CoT, we set the number of training epochs to 2 and use the average of the FKL loss and the CE loss. For pruning, we use the 0.5B-Instruct model as the base, implementing width/depth pruning strategies with a 20\%/30\% prune ratio, resulting in two models Qwen2.5-0.4B(0.35B)-Width and Qwen2.5-0.4B(0.35B)-Depth. These extremely compressed models are produced through a two-phase training process involving accuracy recovery training and knowledge distillation (KD). Comprehensive training details for the pruned models are available in \autoref{sec:prune_detail}. For quantization, we primarily experiment with three strategies on the 0.5B model trained with AKL: W8A16, W4A16, and W8A8. Both W8A16 and W4A16 employ per-group symmetric int8 weight-only quantization with a \verb|group_size=128|, while activations are in fp16. To improve accuracy, W4A16 use 512 training data samples combined with the GPTQ algorithm \cite{frantar2023gptqaccurateposttrainingquantization} for calibration. In contrast, W8A8 uses FP8 for both weights and activations. For the training frameworks, we utilize Megatron-LM \cite{DBLP:conf/sc/NarayananSCLPKV21} for SFT and KD, while OpenRLHF \cite{DBLP:journals/corr/abs-2405-11143} is employed for RL. Data sampling during training is conducted using vLLM \cite{kwon2023efficient}. In the evaluation phase, we employ SGLang \cite{zheng2024sglangefficientexecutionstructured} to generate data.

\section{Computing resources}\label{sec:compute_resources}

We utilize 8 NVIDIA H20 GPUs for training and data generation processes. For evaluating inference performance, a single NVIDIA L20 GPU is employed. We apply an INTEL XEON PLATINUM 8575C processor with 192 cores for computation, operating at a frequency of 3028.956 MHz and a cache size of 327680 KB. Additionally, we establish a virtual development environment based on Kubernetes, leveraging 64 virtual cores and 512 GB of memory. During the training phase, the RFT process spans approximately 2 to 14 hours for models ranging from 0.5B to 7B in size. Pre-KD logits caching for a 7B model requires about 5 hours, whereas RL for the 3B model extends to approximately 3 days. The KD process for the 0.5B model is completed within about 2 hours, and training each pruned model requires roughly 3.5 days. Moreover, for the LLM prototype and LLM-as-a-Judge, generating data for every 10,000 entries takes approximately 1 hour.

\section{Detailed Analysis for Experimental Results for KD}\label{sec:kd_result_details}

As demonstrated in \autoref{tab:kd_result}, it is clear that balancing FKL and RKL is crucial for our task. The function selection process by the model resembles a classification task, whereas generating parameter names and values for function calls is more akin to a generation task. The parameter \(\mu\) in AKL is precisely used to adjust the effects between classification and generation. Experimental results demonstrate that at \(\mu \leq 0.7\) and \(\mu = 1.0\) (equivalent to FKL), the performance with AKL is similar to or worse than RFT. Based on our findings where the average probability of the top-1 token from the teacher model exceeds 99\%, we conclude that setting a lower $\mu$ value causes the effect of AKL to excessively shift more towards RKL. As described in the AKL paper, RKL prioritizes learning the tail of the probability distribution, which can impede rapid convergence in scenarios without requiring diversity. However, RKL can bring considerable benefits for the function classification part. Overall, AKL demonstrates greater potential compared to RFT, achieving performance improvements with an appropriately chosen \(\mu\) value.

\section{Inference Details}\label{sec:infer_details}
The inference framework utilizes TensorRT-LLM\cite{tensorrt-llm} v0.16.0 with enabled optimizations including paged attention, prefix cache, CUDA graph size set to 1024, and max token number set to 32768 for inference acceleration.

\section{Pruning Details}\label{sec:prune_detail}

\subsection{Offline One-shot Pruning}\label{sec:offline_prune}

\begin{figure}[H]
\centering
\includegraphics[width=1\linewidth]{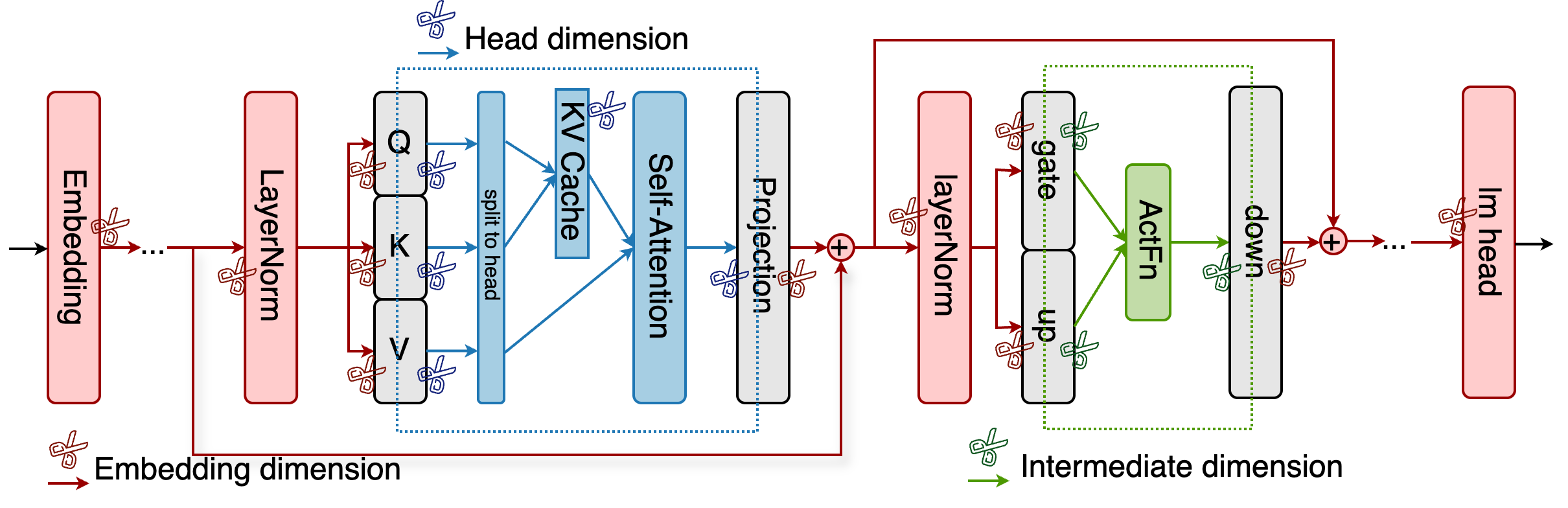}
\caption{\label{fig:prune}Width pruning dependency in the Qwen2.5 model.}
\end{figure}

\paragraph{Importance as Metric for Structured Pruning}
For Depth Pruning, we adopt Layer Importance (LI) as the metric, which is based on the similarity between consecutive layers' activation. Formally, we define the importance of the $i$-th layer as：
    \begin{equation}LI_i = 1 - \mathbb{E}_{X,t} \frac{X_{i,t}^T X_{i+1,t}}{\|X_{i,t}\|_2 \|X_{i+1,t}\|_2}
    \end{equation}
where $X_i$ denotes the activation matrix of the $i$-th layer and $X_{i,t}$ denotes the $t$-th column (embedding vector) of $X_i$. After all layer activations $\{X_i\}_{i=1}^L$ through a single forward pass on calibration data are collected, the $N$ least important layers are removed, where $N$ is determined by the target pruning ratio. For Width Pruning, we classify width pruning into three dimensions based on channel dependencies: Embedding dimension, Head dimension and Intermediate dimension (as illustrated in \autoref{fig:prune}). For compact LLMs (e.g., Qwen2.5-0.5B with head size of 14), we exclude head dimension pruning due to its demonstrated sensitivity\cite{muralidharan2024compactlanguagemodelspruning}. Instead, we employ an activation-based importance metric, Channel Importance(CI), for channel selection as follows：
    \begin{equation}
    CI^{i}_{\text{embedding}} = \sqrt{\sum_{b,s} |\text{LN}(X)^i|^2}, \quad
    CI^{i}_{\text{inter}} = \sqrt{\sum_{b,s} |x (W_{ffn1}^{i})^T|^2}
    \end{equation}

where $CI^{i}_{\text{embedding}}$ denotes the importance score of the $i$-th embedding dimension, computed from the LayerNorm outputs of attention and FFN blocks, while $CI^{i}_{\text{inter}}$ denotes the importance of the $i$-th intermediate dimension, calculated from the $i$-th row of FFN's gate and up projection matrices $W_{ffn1}^{i}$. The importance scores are first aggregated across batch and sequence dimensions through summation $\sum_{b,s}$, then normalized using L2 normalization within each subgroup to obtain global channel importance.

\noindent

\paragraph{Optimal Width Pruning Configuration}
To ensure hardware-efficient matrix multiplication during inference, we set the pruning granularity to 64 for embedding dimension and 256 for the intermediate dimension. This approach results in multiple width-pruned model configurations, as show in \autoref{tab:width_prune_arch_ablation}. We calculate channel importance scores $CI^{i}_{\text{embedding}}$ and $CI^{j}_{\text{inter}}$ using a calibration dataset consisting of 1024 randomly sampled sequences from WikiText-2. For each configuration, we prune entire channel groups with the lowest CI scores and evaluate the resulting one-shot pruned models through WikiText-2 perplexity. \autoref{tab:width_prune_arch_ablation} presents the optimal configurations for width pruning at ratio of 20\% and 30\%, identified as Qwen2.5-0.4B-Width-2 and Qwen2.5-0.35B-Width-3, respectively.

\begin{table}[]
\caption{Perplexity(PPL) evaluation on WikiText-2 comparing model architectures with one-shot width pruning at 20\% and 30\% pruning ratios.}
\label{tab:width_prune_arch_ablation}
\begin{tabular}{l|cccccc}
\toprule
\multicolumn{1}{c|}{Model} & \begin{tabular}[c]{@{}c@{}}Hidden\\ Size\end{tabular} & \begin{tabular}[c]{@{}c@{}}FFN\\ Inter Size\end{tabular} & \begin{tabular}[c]{@{}c@{}}Non-Emb\\ Params(B)\end{tabular} & \begin{tabular}[c]{@{}c@{}}Emb\\ Param(B)\end{tabular} & \begin{tabular}[c]{@{}c@{}}Params\\ (B)\end{tabular} & \begin{tabular}[c]{@{}c@{}}WikiText2\\ PPL\end{tabular} \\ \midrule
Qwen2.5-0.5B               & 896                                                   & 4864                                                     & 0.358                                                       & 0.136                                                  & 0.345                                                & 14.28                                                   \\ \midrule
Qwen2.5-0.4B-Width-1       & 896                                                   & 3328                                                     & 0.259                                                       & 0.136                                                  & 0.395                                                & 349.89                                                  \\
Qwen2.5-0.4B-Width-2       & 832                                                   & 3840                                                     & 0.271                                                       & 0.126                                                  & 0.397                                                & \textbf{251.95}                                         \\
Qwen2.5-0.4B-Width-3       & 768                                                   & 4352                                                     & 0.278                                                       & 0.117                                                  & 0.395                                                & 354.74                                                  \\
Qwen2.5-0.4B-Width-4       & 706                                                   & 4864                                                     & 0.282                                                       & 0.107                                                  & 0.389                                                & 493.55                                                  \\ \midrule
Qwen2.5-0.35B-Width-1      & 896                                                   & 2560                                                     & 0.209                                                       & 0.136                                                  & 0.345                                                & 1210.42                                                 \\
Qwen2.5-0.35B-Width-2      & 768                                                   & 3584                                                     & 0.236                                                       & 0.117                                                  & 0.351                                                & 1141.69                                                 \\
Qwen2.5-0.35B-Width-3      & 768                                                   & 3584                                                     & 0.236                                                       & 0.117                                                  & 0.353                                                & \textbf{1009.57}                                        \\
Qwen2.5-0.35B-Width-4      & 706                                                   & 4096                                                     & 0.242                                                       & 0.107                                                  & 0.349                                                & 2026.00                                                 \\
Qwen2.5-0.35B-Width-5      & 640                                                   & 4864                                                     & 0.256                                                       & 0.097                                                  & 0.353                                                & 2395.52                                                 \\ \bottomrule
\end{tabular}
\end{table}

\subsection{Recovery Training in Pruning}\label{sec:prune_kd_detail}

\noindent
\paragraph{Recovery Training Dataset and Hyperparameters}
We use the Infinity-Instruct-7M\cite{InfinityInstruct2024} dataset and apply AKL distillation\cite{DBLP:conf/coling/WuTWY0W25} described in \autoref{sec:distill} for general instruction recovery training. Qwen2.5-7B-Instruct\cite{qwen2025qwen25technicalreport} serves as the teacher model, and we follow Minitron's methodology\cite{sreenivas2024llmpruningdistillationpractice} to perform teacher correction on the Infinity-Instruct-7M dataset. The Adam optimizer is used with a weight decay of 1e-4 in conjunction with a cosine learning rate schedule, decaying from 2e-5 to 2e-6. We conduct training over 2 epochs, utilizing a batch size of 32, with each batch packed to accommodate sequences of up to 7168 tokens.

\paragraph{Performance across Knowledge Assessment \& Reasoning Tasks}
\autoref{tab:pruned_model_evalation} summarizes the performance of pruned models across various downstream tasks, including MMLU\cite{hendrycks2021measuringmassivemultitasklanguage} and CMMLU\cite{li2024cmmlumeasuringmassivemultitask} for knowledge assessment, alongside ARC-Challenge\cite{clark2018thinksolvedquestionanswering} and HellaSwag\cite{zellers2019hellaswagmachinereallyfinish} for commonsense reasoning evaluation, and GSM8k\cite{cobbe2021trainingverifierssolvemath} for logical reasoning. Our experiments reveal distinct recovery patterns between width and depth pruning across general tasks. At a 20\% pruning rate, width pruning achieves 90.5\% of the performance relative to the unpruned baseline, while depth pruning retains only 78.4\%. When increasing the pruning intensity to 30\%, width pruning preserves 83.8\% of the original performance. In contrast, depth pruning suffers catastrophic performance degradation, particularly on the GSM8K task, where accuracy drops from 45.11 to a merely 5.61.

\begin{table}[]
\centering
\caption{Performance comparison of Qwen2.5 models after pruning. We report 5-shot performance on MMLU and CMMLU, 4-shot for GSM8k, zero-shot for ARC-Challenge, HellaSwag. * denotes that the baseline Qwen-2.5-0.5B utilizes the same instruct fine-tuning dataset for correction fine-tuning.}
\label{tab:pruned_model_evalation}
\begin{tabular}{@{}lcccccl@{}}
\toprule
\multicolumn{1}{c}{Model} & \begin{tabular}[c]{@{}c@{}}MMLU\\ 5-shot\end{tabular} & \begin{tabular}[c]{@{}c@{}}CMMLU\\ 5-shot\end{tabular} & \begin{tabular}[c]{@{}c@{}}ARC-C\\ 0-shot\end{tabular} & \begin{tabular}[c]{@{}c@{}}HellaSwag\\ 0-shot\end{tabular} & \begin{tabular}[c]{@{}c@{}}GSM8k\\ 4-shot\end{tabular} & Avg.  \\ \midrule
Qwen-2.5-0.5B*  & 44.65                                                 & 42.53                                                  & 50.17                                                  & 39.13                                                      & 45.11                                                  & 44.32 \\ \midrule
Qwen-2.5-0.4B-Width       & 38.63                                                 & 35.89                                                  & 42.27                                                  & 35.55                                                      & 48.60                                                  & 40.12 \\
Qwen-2.5-0.35B-Width      & 35.85                                                 & 33.18                                                  & 40.21                                                  & 32.21                                                      & 44.35                                                  & 37.16 \\ \midrule
Qwen-2.5-0.4B-Depth       & 32.89                                                 & 32.41                                                  & 37.11                                                  & 33.69                                                      & 37.68                                                  & 34.76 \\
Qwen-2.5-0.35B-Depth      & 25.46                                                 & 24.42                                                  & 26.12                                                  & 25.21                                                      & 5.61                                                   & 21.36 \\ \bottomrule
\end{tabular}
\end{table}

\noindent
\begin{minipage}{0.56\linewidth}

\paragraph{Ablation of Pruning Ratio on Domain-Specific Task}
Our investigation reveals a nonlinear relationship between pruning intensity and model accuracy in domain-specific tasks. As shown in \autoref{tab:prune_ratio_ablation}, both depth-wise and width-wise pruning at a 20\% pruning rate result in a negligible <0.5\% AR reduction. However, increasing the pruning rate to 30\% leads to notable performance deterioration (>1.5\% AR drop). These findings suggest that for small-scale models on domain-specific tasks, pruning rates exceeding 20\% may significantly compromise accuracy.

\end{minipage}
\hfill
\begin{minipage}{0.41\linewidth}
\vspace{-5pt}
\centering
\captionof{table}{Results in different pruning strategies on domain-specific tasks.}
\vspace{\abovecaptionskip}
\label{tab:prune_ratio_ablation}
\begin{tabular}{@{}lc@{}}
\toprule
\multicolumn{1}{c}{Model} & AR      \\ \midrule
Qwen2.5-0.5B (KD)         & 94.85\% \\ \midrule
Qwen2.5-0.4B-Width (KD)   & 94.17\% \\
Qwen2.5-0.35B-Width (KD)  & 93.71\%       \\ \midrule
Qwen2.5-0.4B-Depth (KD)   & 94.20\% \\
Qwen2.5-0.35B-Depth (KD)  & 93.75\%       \\ \bottomrule
\end{tabular}
\end{minipage}

\section{FLOPs Calculation}\label{sec:FLOPs}

For the calculation of FLOPs, we first define the notations symbolically:
\begin{itemize}
\setlength{\itemsep}{0pt}
\setlength{\parskip}{0pt}
\item $b$: the batch size
\item $h$: the hidden size
\item $s_{\text{t}}$: the total average number of prompt tokens
\item $s_{\text{u}}$: the average number of uncached tokens in prefill phase
\item $n_{\text{a}}$: the number of attention heads
\item $n_{\text{kv}}$: the number of separate key/value heads in GQA
\item $d_{\text{h}}$: the dimension of each head
\item $d_{\text{i}}$: the intermediate dimension of FFN layer
\item $v$: the vocabulary size
\item $l$: the number of model layers
\end{itemize}

According to \cite{korthikanti2022reducingactivationrecomputationlarge}, we calculate the floating-point operations (FLOPs) of the Qwen2.5 models, only considering matrix multiplications (GEMMs) without norms and activations:

\begin{itemize}
\setlength{\itemsep}{0pt}
\setlength{\parskip}{0pt}
    \item \textbf{Attention block operations:}
    \begin{itemize}
        \item Query/Key/Value transformations (GQA): $2bs_{\text{u}}hd_{\text{h}}(n_{\text{a}} + 2n_{\text{kv}})$
        \item Attention matrix computation and attention over values: $2bs_{\text{t}}s_{\text{u}}n_{\text{a}}d_{\text{h}}$
        \item Post-attention linear projection: $2bs_{\text{u}}hn_{\text{a}}d_{\text{h}}$
    \end{itemize}

    \item \textbf{Feed-forward network block operations:}
    \begin{itemize}
        \item Up/Gate/Down projections (each): $2bs_{\text{u}}hd_{\text{i}}$
    \end{itemize}

    \item \textbf{LM head operations:} $2bhv$

\end{itemize}

Thus, the total FLOPs for one inference with prefill cache is:

\begin{align*}
\text{FLOPs}_{\ \text{total}} &= l \Bigg[4bs_{\text{u}} h d_{\text{h}} \left(n_{\text{a}} + n_{\text{kv}}\right) + 2bs_{\text{u}}s_{\text{t}} d_{\text{h}} n_{\text{a}} + 6bs_{\text{u}}h d_{\text{i}} \Bigg] + 2bhv
\end{align*}

\paragraph{Workload and Computational Analysis}
We conduct experiments on our datasets, in which prompts have an average length of 1792 tokens. By leveraging prefix caching to exploit shared prompt prefixes during inference, we reduce the average number of uncached tokens in the prefill phase to 128 and generate an average of 9 tokens per request. We calculate FLOPs under $s_{\text{u}} = 128$ tokens in the prefill phase and under $s_{\text{u}} = 1$ token for each step in the decoding phase, with detailed results presented in \autoref{tab:main_result}.

\section{Extensive Attempts on Generalization Enhancement}\label{sec:full_result}
\begin{table}[!h]
\centering
\caption{In-domain and out-of-domain performance of Qwen2.5 models using different knowledge transfer strategies.}
\label{tab:full_results}
\begin{tabular}{l|r|r}
\toprule
Model & \multicolumn{1}{p{2cm}}{\centering AR \\ (In-Domain)} & \multicolumn{1}{|p{2.5cm}}{\centering AR \\ (Out-of-Domain)} \\
\midrule

BERT-based system & 81.43\% & - \\
Qwen2.5-72B (ICL) & 94.74\% & \underline{94.63}\% \\
Qwen2.5-7B (RFT) & 95.29\% & 90.86\% \\
Qwen2.5-3B (RFT) & 95.30\% & 91.02\% \\
Qwen2.5-0.5B (RFT) & 94.66\% & 90.62\% \\
Qwen2.5-0.5B (KD) & 94.85\% & 90.53\% \\
Qwen2.5-7B (RL) & 95.36\% & \textbf{94.80}\% \\
Qwen2.5-3B (RL) & 94.68\% & 92.94\% \\
Qwen2.5-0.5B (RL) & 90.76\% & 85.21\% \\
\midrule
\multicolumn{3}{c}{\textbf{Knowledge Transfer using Qwen2.5-3B (RL)}} \\
\midrule
Qwen2.5-0.5B (RFT\textsuperscript{*}) & 93.82\% & 89.13\% \\
Qwen2.5-0.5B (RFT\textsuperscript{*}+RL) & 95.72\% & 90.09\% \\
Qwen2.5-0.5B (KD\textsuperscript{*}) & 88.49\% & 84.14\% \\
Qwen2.5-0.5B (KD\textsuperscript{*}+RL) & 95.55\% & 91.68\% \\
\midrule
\multicolumn{3}{c}{\textbf{Knowledge Transfer using Qwen2.5-7B (RL)}} \\
\midrule
Qwen2.5-0.5B (RFT\textsuperscript{*}) & 95.29\% & 90.28\% \\
Qwen2.5-0.5B (RFT\textsuperscript{*}+RL) & \underline{95.94}\% & 92.21\% \\
Qwen2.5-0.5B (KD\textsuperscript{*}) & 95.32\% & 90.59\% \\
Qwen2.5-0.5B (KD\textsuperscript{*}+RL) & \textbf{95.99}\% & 92.34\% \\
\bottomrule
\end{tabular}
\end{table}

\noindent
\begin{figure}
\begin{minipage}{0.43\linewidth}
    \centering
    \includegraphics[width=1.0\linewidth]{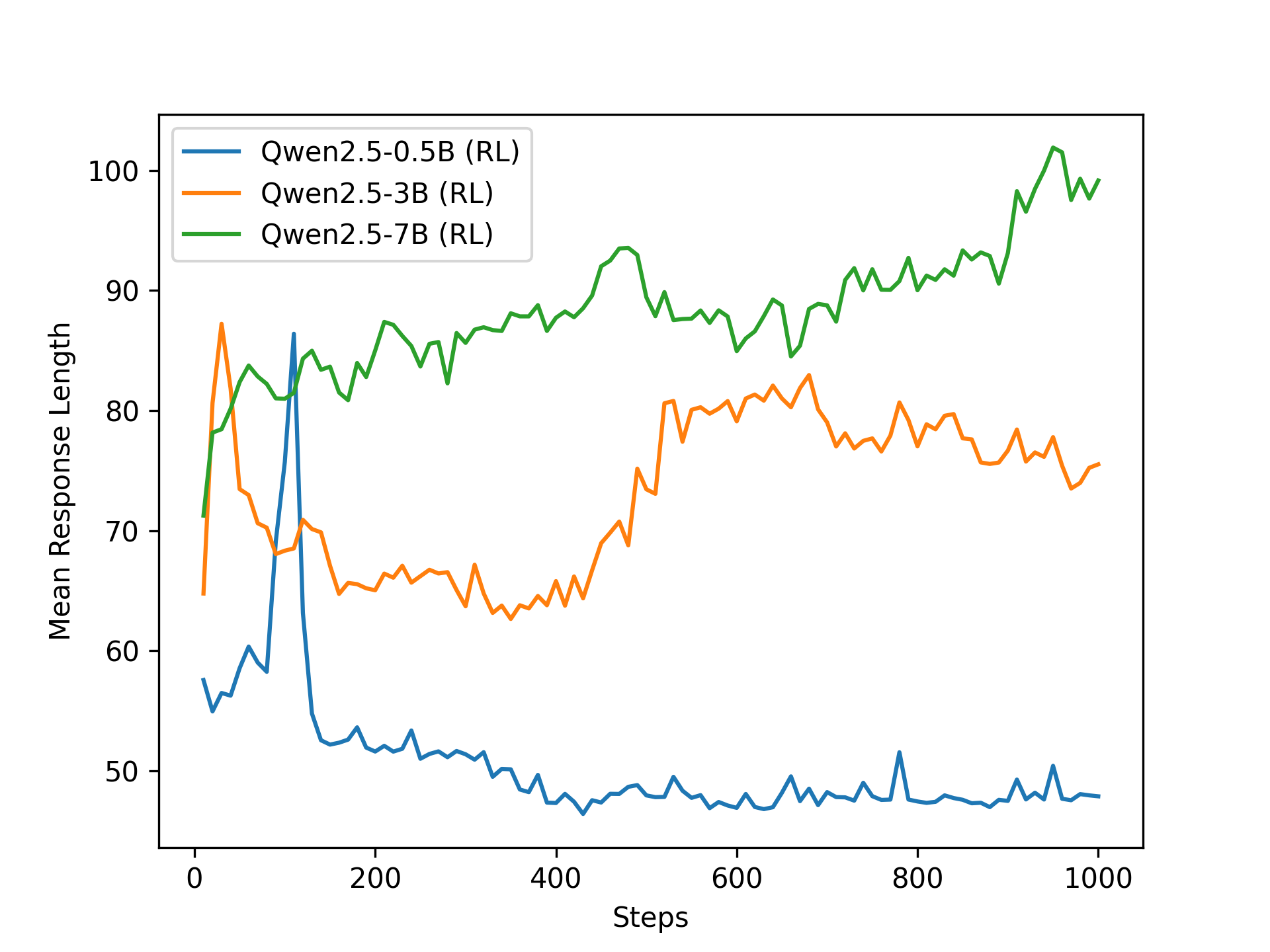}
    \caption{Response lengths of different models during RL training}
    \label{fig:rl_resp_len}
\end{minipage}
\hfill
\begin{minipage}{0.52\linewidth}

    \centering
    \captionof{table}{Response lengths of different models, represented as mean $\pm$ standard deviation.}
    \vspace{\abovecaptionskip}
    \label{tab:resp_lens}
    \begin{tabular}{l|r}
    \toprule
    Model & Response Length \\
    \midrule
    QwQ-32B (ICL) & 316.8 $\pm$ 111.4 \\
    Qwen2.5-7B (RL) & 86.4 $\pm$ 12.9 \\
    Qwen2.5-3B (RL) & 61.7 $\pm$ 9.4 \\
    Qwen2.5-0.5B (KD\textsuperscript{*}) & 86.3 $\pm$ 12.6 \\
    Qwen2.5-0.5B (KD\textsuperscript{*}+RL) & 85.3 $\pm$ 10.8 \\
    \bottomrule
    \end{tabular}
\end{minipage}

\end{figure}

Upon discovering that using RFT or KD alone leads to significant degradation in out-of-domain performance for super-tiny models, we conduct extensive experiments to enhance generalization capabilities, as illustrated in \autoref{tab:full_results}. Initially, we prioritize Reinforcement Learning (RL) for this task. As expected, 7B RL-trained models achieve excellent results, surpassing the LLM Prototype and RFT-trained models. Subsequently, we attempt direct RL training on smaller models, such as 3B and 0.5B. The results are less favorable this time. While 3B RL-trained models exhibit considerable generalization capabilities, 0.5B RL-trained models show no improvement in generalization, underperforming even compared to 0.5B RFT-trained models. Through observations and analyses, we conclude that this is because models of such small size tend to generate low-quality Chain-of-Thought (CoT), as shown in \autoref{fig:rl_resp_len}, leading to degraded output diversity, akin to scenarios where CoT is not used.

To address the challenge of 0.5B models' inabilities to generate high-quality CoT, we propose a hybrid strategy. Given the favorable results from 7B RL-trained models, we decide to use 7B RL-trained models as our starting point. Then, we transfer out-of-domain capabilities and the abilities of generating high-quality CoT from 7B RL-trained models to 0.5B models via Rejection sampling Fine-Tuning (RFT) or knowledge distillation (KD), which are denoted as RFT\textsuperscript{*} and KD\textsuperscript{*}, respectively. Experimental results indicate that this knowledge transfer indeed outperforms direct RL training on 0.5B models in enhancing generalization and preserving the abilities of generating high-quality CoT.

So far, 0.5B KD\textsuperscript{*} models posses the abilities to generate high-quality CoT, but in terms of out-of-domain Achievable Rate (AR), there is no considerable improvement compared to 0.5B RFT-trained models. Consequently, we conduct another round of RL training on 0.5B KD\textsuperscript{*} models with data generated by the LLM prototype for enhancing generalization, similar to how we initially improve 7B models' generalization with RL. Ultimately, experimental results present that after the second RL training, the 0.5B super-tiny models exhibit impressive improvement on model performance, with both an absolute increase of over 1\% AR increase on in-domain and out-of-domain datasets compared to 0.5B RFT-trained models.

\begin{table}
\centering
\caption{Ablation of knowledge transfer strategies on Qwen2.5-0.5B models.}
\label{tab:ablation_kd_ratios}
\begin{tabular}{l|r|r}
\toprule
Model & \multicolumn{1}{p{2cm}}{\centering AR \\ (In-Domain)} & \multicolumn{1}{|p{2.5cm}}{\centering AR \\ (Out-of-Domain)} \\
Qwen2.5-0.5B (RFT\textsuperscript{*}) & 95.29\% & 90.28\% \\
Qwen2.5-0.5B (KD\textsuperscript{*}[FKL]) & 95.08\% & 90.19\% \\
Qwen2.5-0.5B (KD\textsuperscript{*}[0.5FKL+0.5CE]) & 95.32\% & 90.59\% \\
Qwen2.5-0.5B (KD\textsuperscript{*}[AKL]) & 93.93\% & 88.37\% \\
Qwen2.5-0.5B (KD\textsuperscript{*}[0.5AKL+0.5CE]) & 95.82\% & 90.85\% \\
Qwen2.5-0.5B (RFT\textsuperscript{*}+RL) & 95.94\% & \underline{92.21}\% \\
Qwen2.5-0.5B (KD\textsuperscript{*}[FKL]+RL) & 95.66\% & \underline{92.21}\% \\
Qwen2.5-0.5B (KD\textsuperscript{*}[0.5FKL+0.5CE]+RL) & \underline{95.99}\% & \textbf{92.34}\% \\
Qwen2.5-0.5B (KD\textsuperscript{*}[AKL]+RL) & 95.85\% & 92.10\% \\
Qwen2.5-0.5B (KD\textsuperscript{*}[0.5AKL+0.5CE]+RL) & \textbf{96.10}\% & 92.03\% \\
\bottomrule
\end{tabular}
\end{table}

\paragraph{Ablation of Different Teacher Models} As illustrated in \autoref{tab:full_results}, a teacher model with inherently superior capabilities is crucial for enhancing student model performance, particularly regarding out-of-domain capabilities. A better teacher model not only leads to better performance but also ensures greater stability in the student models' results, driven by the generation of higher-quality CoT.

\paragraph{Response Lengths of Different Reasoning Models} As can be seen in \autoref{tab:resp_lens}, the average response lengths of our models are more than three times shorter than that of QwQ-32B. For direct RL training, longer response lengths appear to yield better results. However, this conclusion is not entirely valid. For instance, in the case of KD\textsuperscript{*} models, there is almost no change in response lengths before and after RL, yet there is a significant difference in out-of-domain performance.

\paragraph{Ablation of Knowledge Transfer Strategies} During the knowledge transfer stage from 7B RL-trained Qwen2.5 models, we investigate the effectiveness of different knowledge transfer strategies, as illustrated in \autoref{tab:ablation_kd_ratios}. Apart from the KD\textsuperscript{*}[AKL] strategy, which encounters a certain degree of underfitting, other experimental results demonstrate the robustness of different knowledge transfer strategies. Specifically, utilizing a loss with Cross Entropy (CE) is preferable to not incorporating CE during KD, as the high-quality CoT generated by 7B RL-trained models facilitates faster convergence with CE loss. Conversely, when the teacher model generates low-quality CoT, using CE alone, equivalent to RFT, can lead to overfitting, as evidenced in \autoref{tab:full_results}, where the 3B RL-trained model as the teacher model exhibits poorer performance.

\section{Experiments on Public Datasets}\label{sec:exp_on_pub_datasets}

\begin{table}[htbp]
\centering
\caption{Performance of different models on BFCL and xlam-function-calling-60k datasets.}

\begin{tabular}{lcccc}
\toprule
\multicolumn{1}{c}{} & \multicolumn{3}{c}{BFCL} & \multicolumn{1}{c}{In-domain} \\
\cmidrule(lr){2-4} \cmidrule(lr){5-5}
Model & Overall & Non-live & Live & Acc \\
\midrule
Qwen2.5-0.5B & 55.14 & 62.29 & 47.98 & 55.88 \\
Hammer2.0-0.5B \cite{lin2025robust}     & 59.31 & 67.00 & 51.62 & /     \\
Qwen2.5-72B & 83.55 & 88.13 & 78.98 & 79.47 \\
Qwen2.5-7B (RL)   & 81.85 & 86.27 & 77.42 & 78.55 \\
Qwen2.5-0.5B (RL) & 55.15 & 64.48 & 45.82 & 65.68 \\
Qwen2.5-0.5B (RFT)     & 56.64 & 64.50 & 48.78 & 74.08 \\
Qwen2.5-0.4B-Depth (RFT)  & 42.27	& 54.56	& 29.98	& 71.55 \\
Qwen2.5-0.4B-Width (RFT)  & 42.60	& 56.17	& 29.02	& 72.98 \\
Qwen2.5-0.5B-W8A16 (RFT)  & 56.42	& 64.42	& 48.41	& 74.38 \\
Qwen2.5-0.5B-W4A16 (RFT)  & 49.54	& 58.52	& 40.56	& 72.23 \\
Qwen2.5-0.5B-W8A8  (RFT)  & 55.68	& 62.88	& 48.48	& 74.06 \\
\midrule
\multicolumn{5}{c}{\textbf{Knowledge Transfer using Qwen2.5-7B (RL)}} \\
\midrule
Qwen2.5-0.5B (KD\textsuperscript{*}[FKL])  & 59.44 & 69.06 & 49.81 & 72.97 \\
Qwen2.5-0.5B (KD\textsuperscript{*}[0.5FKL+0.5CE]) & 54.28 & 62.00 & 46.56 & 66.55 \\
Qwen2.5-0.5B (KD\textsuperscript{*}[AKL])  & 58.45 & 67.15 & 49.74 & 71.90 \\
Qwen2.5-0.5B (KD\textsuperscript{*}[0.5AKL+0.5CE])  & 58.97 & 68.12 & 49.81 & 71.70 \\
Qwen2.5-0.5B (KD\textsuperscript{*}[FKL]+RL)  & 60.42 & \textbf{71.02} & 49.81 & 75.87 \\
Qwen2.5-0.5B (KD\textsuperscript{*}[0.5FKL+0.5CE]+RL) & 59.84 & \underline{70.00} & 49.67 & \textbf{77.02} \\
Qwen2.5-0.5B (KD\textsuperscript{*}[AKL]+RL)  & \underline{60.50} & 69.56 & \underline{51.44} & 76.38 \\
Qwen2.5-0.5B (KD\textsuperscript{*}[0.5AKL+0.5CE]+RL)  & \textbf{60.79} & 69.46 & \textbf{52.11} & \underline{76.68} \\
\bottomrule
\end{tabular}
\label{tab:public_experiments}
\end{table}

To validate the effectiveness of our proposed method, we conduct experiments on publicly available datasets. We randomly partition 90\% of the xlam-function-calling-60k dataset \cite{DBLP:conf/nips/LiuHZZLKTYLFNYS24} for training, with the remaining 10\% serving as an in-domain test set. For the training set, instead of utilizing the provided labels, we employ Qwen2.5-72B-Instruct \cite{qwen2025qwen25technicalreport} to generate reference labels. Furthermore, we design a judge based on QwQ-32B \cite{qwq32b} to assess the accuracy of the labels generated by the LLM. This process can be viewed as prototyping. Given that function calling is a standardized task, we use the official function call prompt template instead of developing a specialized agent for answer generation. The design of our judge is inspired by the Format Checker and Semantic Checker proposed in \cite{DBLP:conf/nips/LiuHZZLKTYLFNYS24}. Through this procedure, we ultimately curate 50,201 valid query-tools-answers pairs from the initial 54k training samples for subsequent training.

Due to the complex function call formats within this dataset, we redesign the reward mechanism for Reinforcement Learning (RL). The reward is defined as:
\begin{equation}
    Reward = \alpha \cdot Reward_{format} + \beta \cdot Reward_{answer}
\end{equation}
where $Reward_{format}$ is 1 if and only if all the following conditions are met, and 0 otherwise: (1) The model's output contains exactly one "<think>...</think>" block; (2) The function call within the answer can be parsed by the official Qwen prompt template and can be loaded as a JSON object; (3) The function name in the answer exists within the tools provided in the prompt, and the parameter types conform to their definitions, which is implemented using Pydantic\cite{colvin2025pydantic}.

For $Reward_{answer}$, we first check whether the number of functions in the model's output is equivalent to that of the ground truth: If not, $Reward_{answer}$ is 0; Otherwise, $Reward_{answer}$ is defined as follows:
\begin{equation}
Reward_i =
\begin{cases}
\frac{\lvert P_{i, pred} \cap P_{i, label} \rvert}{\lvert P_{i, pred} \cup P_{i, label}\rvert}, & \text{if }\lvert P_{i, pred} \cup P_{i, label}\rvert > 0 \\
1, & \text{otherwise}
\end{cases}
\end{equation}
\begin{equation}
    Reward_{answer} = \frac{1}{N}\sum Reward_i
\end{equation}
where $N$ is the number of function calls, and $P_i$ denotes the set of parameters for the $i$-th function call. During training, we set $\alpha=1$ and $\beta=2$. The baseline variant of Reinforce++ \cite{hu2025reinforcesimpleefficientapproach} is selected as our reinforcement learning method.

We evaluate the trained model on both in-domain and out-of-domain dataset. For in-domain evaluation, we utilize the aforementioned test set with official answers as the ground truth. A test instance is considered correct only if the model's output exactly matches the answer (i.e., identical function names and key-value pairs for arguments); all other cases are deemed incorrect. Accuracy is calculated based on this criterion. Additionally, we employ the Berkeley Function Call Leaderboard (BFCL)\cite{berkeley-function-calling-leaderboard} for out-of-domain evaluation. The final experimental results are presented in \autoref{tab:public_experiments}.

The experimental results indicate that the 0.5B trained model achieves in-domain accuracy comparable to the Qwen2.5-72B model. Furthermore, it demonstrates a substantial improvement over Qwen2.5-0.5B-Instruct in out-of-domain scenarios. We also compare our method with Hammer2.0-0.5B\cite{lin2025robust}, the previous SOTA 0.5B model, which is also fine-tuned from Qwen2.5-0.5B-Instruct8. Our approach achieves superior performance, demonstrating the effectiveness of our proposed method. In terms of quantization, experimental results demonstrate that applying W8A8 and W8A16 strategies results in negligible loss compared to the original model, while using the W4A16 strategy leads to significant loss in out-of-domain scenarios. Regarding pruning, all pruned models exhibit substantial losses in out-of-domain scenarios, yet own relatively minor loss in in-domain scenarios.

\end{CJK}

\begin{thebibliography}{10}

\bibitem{guo2021selfattentionexternalattentionusing}
Meng-Hao Guo, Zheng-Ning Liu, Tai-Jiang Mu, and Shi-Min Hu.
\newblock Beyond self-attention: External attention using two linear layers for
  visual tasks, 2021.

\bibitem{thoppilan2022lamdalanguagemodelsdialog}
Romal Thoppilan, Daniel~De Freitas, Jamie Hall, Noam Shazeer, Apoorv
  Kulshreshtha, Heng-Tze Cheng, Alicia Jin, Taylor Bos, Leslie Baker, Yu~Du,
  YaGuang Li, Hongrae Lee, Huaixiu~Steven Zheng, Amin Ghafouri, Marcelo
  Menegali, Yanping Huang, Maxim Krikun, Dmitry Lepikhin, James Qin, Dehao
  Chen, Yuanzhong Xu, Zhifeng Chen, Adam Roberts, Maarten Bosma, Vincent Zhao,
  Yanqi Zhou, Chung-Ching Chang, Igor Krivokon, Will Rusch, Marc Pickett,
  Pranesh Srinivasan, Laichee Man, Kathleen Meier-Hellstern, Meredith~Ringel
  Morris, Tulsee Doshi, Renelito~Delos Santos, Toju Duke, Johnny Soraker, Ben
  Zevenbergen, Vinodkumar Prabhakaran, Mark Diaz, Ben Hutchinson, Kristen
  Olson, Alejandra Molina, Erin Hoffman-John, Josh Lee, Lora Aroyo, Ravi
  Rajakumar, Alena Butryna, Matthew Lamm, Viktoriya Kuzmina, Joe Fenton, Aaron
  Cohen, Rachel Bernstein, Ray Kurzweil, Blaise Aguera-Arcas, Claire Cui,
  Marian Croak, Ed~Chi, and Quoc Le.
\newblock Lamda: Language models for dialog applications, 2022.

\bibitem{akbiketal2018contextual}
Alan Akbik, Duncan Blythe, and Roland Vollgraf.
\newblock Contextual string embeddings for sequence labeling.
\newblock In Emily~M. Bender, Leon Derczynski, and Pierre Isabelle, editors,
  {\em Proceedings of the 27th International Conference on Computational
  Linguistics}, pages 1638--1649, Santa Fe, New Mexico, USA, August 2018.
  Association for Computational Linguistics.

\bibitem{narayanan2021efficientlargescalelanguagemodel}
Deepak Narayanan, Mohammad Shoeybi, Jared Casper, Patrick LeGresley, Mostofa
  Patwary, Vijay~Anand Korthikanti, Dmitri Vainbrand, Prethvi Kashinkunti,
  Julie Bernauer, Bryan Catanzaro, Amar Phanishayee, and Matei Zaharia.
\newblock Efficient large-scale language model training on gpu clusters using
  megatron-lm, 2021.

\bibitem{crossdatasetposeestimation}
Mo~Zhao, Ya~Ma, Zhendong Li, and Hao Liu.
\newblock Cross-dataset pose estimation of faces in the wild.
\newblock In {\em 2021 5th Asian Conference on Artificial Intelligence
  Technology (ACAIT)}, pages 718--724, 2021.

\bibitem{gusak2022surveylargescaleneural}
Julia Gusak, Daria Cherniuk, Alena Shilova, Alexander Katrutsa, Daniel
  Bershatsky, Xunyi Zhao, Lionel Eyraud-Dubois, Oleg Shlyazhko, Denis Dimitrov,
  Ivan Oseledets, and Olivier Beaumont.
\newblock Survey on large scale neural network training, 2022.

\bibitem{li2025midasmultilevelintentdomain}
Yan Li, So-Eon Kim, Seong-Bae Park, and Soyeon~Caren Han.
\newblock Midas: Multi-level intent, domain, and slot knowledge distillation
  for multi-turn nlu, 2025.

\bibitem{OpenReasonerZero2025}
Jingcheng Hu, Yinmin Zhang, Qi~Han, Daxin Jiang, and Heung-Yeung~Shum
  Xiangyu~Zhang.
\newblock Open-reasoner-zero: An open source approach to scaling reinforcement
  learning on the base model.
\newblock \url{https://github.com/Open-Reasoner-Zero/Open-Reasoner-Zero}, 2025.

\bibitem{xie2025logicrlunleashingllmreasoning}
Tian Xie, Zitian Gao, Qingnan Ren, Haoming Luo, Yuqian Hong, Bryan Dai, Joey
  Zhou, Kai Qiu, Zhirong Wu, and Chong Luo.
\newblock Logic-rl: Unleashing llm reasoning with rule-based reinforcement
  learning, 2025.

\bibitem{tinyzero}
Jiayi Pan, Junjie Zhang, Xingyao Wang, Lifan Yuan, Hao Peng, and Alane Suhr.
\newblock Tinyzero.
\newblock https://github.com/Jiayi-Pan/TinyZero, 2025.
\newblock Accessed: 2025-01-24.

\bibitem{qwen2025qwen25technicalreport}
Qwen, :, An~Yang, Baosong Yang, Beichen Zhang, Binyuan Hui, Bo~Zheng, Bowen Yu,
  Chengyuan Li, Dayiheng Liu, Fei Huang, Haoran Wei, Huan Lin, Jian Yang,
  Jianhong Tu, Jianwei Zhang, Jianxin Yang, Jiaxi Yang, Jingren Zhou, Junyang
  Lin, Kai Dang, Keming Lu, Keqin Bao, Kexin Yang, Le~Yu, Mei Li, Mingfeng Xue,
  Pei Zhang, Qin Zhu, Rui Men, Runji Lin, Tianhao Li, Tianyi Tang, Tingyu Xia,
  Xingzhang Ren, Xuancheng Ren, Yang Fan, Yang Su, Yichang Zhang, Yu~Wan,
  Yuqiong Liu, Zeyu Cui, Zhenru Zhang, and Zihan Qiu.
\newblock Qwen2.5 technical report, 2025.

\bibitem{DBLP:conf/nips/SchickDDRLHZCS23}
Timo Schick, Jane Dwivedi{-}Yu, Roberto Dess{\`{\i}}, Roberta Raileanu, Maria
  Lomeli, Eric Hambro, Luke Zettlemoyer, Nicola Cancedda, and Thomas Scialom.
\newblock Toolformer: Language models can teach themselves to use tools.
\newblock In Alice Oh, Tristan Naumann, Amir Globerson, Kate Saenko, Moritz
  Hardt, and Sergey Levine, editors, {\em Advances in Neural Information
  Processing Systems 36: Annual Conference on Neural Information Processing
  Systems 2023, NeurIPS 2023, New Orleans, LA, USA, December 10 - 16, 2023},
  2023.

\bibitem{DBLP:conf/nips/PatilZ0G24}
Shishir~G. Patil, Tianjun Zhang, Xin Wang, and Joseph~E. Gonzalez.
\newblock Gorilla: Large language model connected with massive apis.
\newblock In Amir Globersons, Lester Mackey, Danielle Belgrave, Angela Fan,
  Ulrich Paquet, Jakub~M. Tomczak, and Cheng Zhang, editors, {\em Advances in
  Neural Information Processing Systems 38: Annual Conference on Neural
  Information Processing Systems 2024, NeurIPS 2024, Vancouver, BC, Canada,
  December 10 - 15, 2024}, 2024.

\bibitem{erdogan2024tinyagentfunctioncallingedge}
Lutfi~Eren Erdogan, Nicholas Lee, Siddharth Jha, Sehoon Kim, Ryan Tabrizi,
  Suhong Moon, Coleman Hooper, Gopala Anumanchipalli, Kurt Keutzer, and Amir
  Gholami.
\newblock Tinyagent: Function calling at the edge, 2024.

\bibitem{kachuee2024improvingtoolretrievalleveraging}
Mohammad Kachuee, Sarthak Ahuja, Vaibhav Kumar, Puyang Xu, and Xiaohu Liu.
\newblock Improving tool retrieval by leveraging large language models for
  query generation, 2024.

\bibitem{DBLP:conf/eamt/KocmiF23}
Tom Kocmi and Christian Federmann.
\newblock Large language models are state-of-the-art evaluators of translation
  quality.
\newblock In Mary Nurminen, Judith Brenner, Maarit Koponen, Sirkku Latomaa,
  Mikhail Mikhailov, Frederike Schierl, Tharindu Ranasinghe, Eva Vanmassenhove,
  Sergi~Alvarez Vidal, Nora Aranberri, Mara Nunziatini, Carla~Parra
  Escart{\'{\i}}n, Mikel~L. Forcada, Maja Popovic, Carolina Scarton, and Helena
  Moniz, editors, {\em Proceedings of the 24th Annual Conference of the
  European Association for Machine Translation, {EAMT} 2023, Tampere, Finland,
  12-15 June 2023}, pages 193--203. European Association for Machine
  Translation, 2023.

\bibitem{DBLP:conf/nips/ZhengC00WZL0LXZ23}
Lianmin Zheng, Wei{-}Lin Chiang, Ying Sheng, Siyuan Zhuang, Zhanghao Wu,
  Yonghao Zhuang, Zi~Lin, Zhuohan Li, Dacheng Li, Eric~P. Xing, Hao Zhang,
  Joseph~E. Gonzalez, and Ion Stoica.
\newblock Judging llm-as-a-judge with mt-bench and chatbot arena.
\newblock In Alice Oh, Tristan Naumann, Amir Globerson, Kate Saenko, Moritz
  Hardt, and Sergey Levine, editors, {\em Advances in Neural Information
  Processing Systems 36: Annual Conference on Neural Information Processing
  Systems 2023, NeurIPS 2023, New Orleans, LA, USA, December 10 - 16, 2023},
  2023.

\bibitem{mcaleese2024llmcriticshelpcatch}
Nat McAleese, Rai~Michael Pokorny, Juan Felipe~Ceron Uribe, Evgenia
  Nitishinskaya, Maja Trebacz, and Jan Leike.
\newblock Llm critics help catch llm bugs, 2024.

\bibitem{DBLP:journals/corr/HintonVD15}
Geoffrey~E. Hinton, Oriol Vinyals, and Jeffrey Dean.
\newblock Distilling the knowledge in a neural network.
\newblock {\em CoRR}, abs/1503.02531, 2015.

\bibitem{DBLP:conf/iclr/Gu0WH24}
Yuxian Gu, Li~Dong, Furu Wei, and Minlie Huang.
\newblock Minillm: Knowledge distillation of large language models.
\newblock In {\em The Twelfth International Conference on Learning
  Representations, {ICLR} 2024, Vienna, Austria, May 7-11, 2024}.
  OpenReview.net, 2024.

\bibitem{dpkd}
Yixing Li, Yuxian Gu, Li~Dong, Dequan Wang, Yu~Cheng, and Furu Wei.
\newblock Direct preference knowledge distillation for large language models.
\newblock {\em arXiv preprint arXiv:2406.19774}, 2024.

\bibitem{yang2023categoriesresponsebasedfeaturebasedrelationbased}
Chuanguang Yang, Xinqiang Yu, Zhulin An, and Yongjun Xu.
\newblock Categories of response-based, feature-based, and relation-based
  knowledge distillation, 2023.

\bibitem{DBLP:conf/emnlp/KimR16}
Yoon Kim and Alexander~M. Rush.
\newblock Sequence-level knowledge distillation.
\newblock In Jian Su, Xavier Carreras, and Kevin Duh, editors, {\em Proceedings
  of the 2016 Conference on Empirical Methods in Natural Language Processing,
  {EMNLP} 2016, Austin, Texas, USA, November 1-4, 2016}, pages 1317--1327. The
  Association for Computational Linguistics, 2016.

\bibitem{kim2023tokenscaledlogitdistillationternary}
Minsoo Kim, Sihwa Lee, Janghwan Lee, Sukjin Hong, Du-Seong Chang, Wonyong Sung,
  and Jungwook Choi.
\newblock Token-scaled logit distillation for ternary weight generative
  language models, 2023.

\bibitem{sanh2020distilbertdistilledversionbert}
Victor Sanh, Lysandre Debut, Julien Chaumond, and Thomas Wolf.
\newblock Distilbert, a distilled version of bert: smaller, faster, cheaper and
  lighter, 2020.

\bibitem{gu2024minillmknowledgedistillationlarge}
Yuxian Gu, Li~Dong, Furu Wei, and Minlie Huang.
\newblock Minillm: Knowledge distillation of large language models, 2024.

\bibitem{agarwal2024onpolicydistillationlanguagemodels}
Rishabh Agarwal, Nino Vieillard, Yongchao Zhou, Piotr Stanczyk, Sabela Ramos,
  Matthieu Geist, and Olivier Bachem.
\newblock On-policy distillation of language models: Learning from
  self-generated mistakes, 2024.

\bibitem{kim2024promptkddistillingstudentfriendlyknowledge}
Gyeongman Kim, Doohyuk Jang, and Eunho Yang.
\newblock Promptkd: Distilling student-friendly knowledge for generative
  language models via prompt tuning, 2024.

\bibitem{DBLP:conf/coling/WuTWY0W25}
Taiqiang Wu, Chaofan Tao, Jiahao Wang, Runming Yang, Zhe Zhao, and Ngai Wong.
\newblock Rethinking kullback-leibler divergence in knowledge distillation for
  large language models.
\newblock In Owen Rambow, Leo Wanner, Marianna Apidianaki, Hend Al{-}Khalifa,
  Barbara~Di Eugenio, and Steven Schockaert, editors, {\em Proceedings of the
  31st International Conference on Computational Linguistics, {COLING} 2025,
  Abu Dhabi, UAE, January 19-24, 2025}, pages 5737--5755. Association for
  Computational Linguistics, 2025.

\bibitem{frantar2023sparsegptmassivelanguagemodels}
Elias Frantar and Dan Alistarh.
\newblock Sparsegpt: Massive language models can be accurately pruned in
  one-shot, 2023.

\bibitem{ma2023llmprunerstructuralpruninglarge}
Xinyin Ma, Gongfan Fang, and Xinchao Wang.
\newblock Llm-pruner: On the structural pruning of large language models, 2023.

\bibitem{dettmers2022llmint88bitmatrixmultiplication}
Tim Dettmers, Mike Lewis, Younes Belkada, and Luke Zettlemoyer.
\newblock Llm.int8(): 8-bit matrix multiplication for transformers at scale,
  2022.

\bibitem{frantar2023gptqaccurateposttrainingquantization}
Elias Frantar, Saleh Ashkboos, Torsten Hoefler, and Dan Alistarh.
\newblock Gptq: Accurate post-training quantization for generative pre-trained
  transformers, 2023.

\bibitem{xiao2024smoothquantaccurateefficientposttraining}
Guangxuan Xiao, Ji~Lin, Mickael Seznec, Hao Wu, Julien Demouth, and Song Han.
\newblock Smoothquant: Accurate and efficient post-training quantization for
  large language models, 2024.

\bibitem{li2017pruningfiltersefficientconvnets}
Hao Li, Asim Kadav, Igor Durdanovic, Hanan Samet, and Hans~Peter Graf.
\newblock Pruning filters for efficient convnets, 2017.

\bibitem{men2024shortgptlayerslargelanguage}
Xin Men, Mingyu Xu, Qingyu Zhang, Bingning Wang, Hongyu Lin, Yaojie Lu, Xianpei
  Han, and Weipeng Chen.
\newblock Shortgpt: Layers in large language models are more redundant than you
  expect, 2024.

\bibitem{muralidharan2024compactlanguagemodelspruning}
Saurav Muralidharan, Sharath~Turuvekere Sreenivas, Raviraj Joshi, Marcin
  Chochowski, Mostofa Patwary, Mohammad Shoeybi, Bryan Catanzaro, Jan Kautz,
  and Pavlo Molchanov.
\newblock Compact language models via pruning and knowledge distillation, 2024.

\bibitem{xia2024shearedllamaacceleratinglanguage}
Mengzhou Xia, Tianyu Gao, Zhiyuan Zeng, and Danqi Chen.
\newblock Sheared llama: Accelerating language model pre-training via
  structured pruning, 2024.

\bibitem{guo2023compressostructuredpruningcollaborative}
Song Guo, Jiahang Xu, Li~Lyna Zhang, and Mao Yang.
\newblock Compresso: Structured pruning with collaborative prompting learns
  compact large language models, 2023.

\bibitem{lin2024awqactivationawareweightquantization}
Ji~Lin, Jiaming Tang, Haotian Tang, Shang Yang, Wei-Ming Chen, Wei-Chen Wang,
  Guangxuan Xiao, Xingyu Dang, Chuang Gan, and Song Han.
\newblock Awq: Activation-aware weight quantization for llm compression and
  acceleration, 2024.

\bibitem{kuzmin2024fp8quantizationpowerexponent}
Andrey Kuzmin, Mart~Van Baalen, Yuwei Ren, Markus Nagel, Jorn Peters, and
  Tijmen Blankevoort.
\newblock Fp8 quantization: The power of the exponent, 2024.

\bibitem{lin2024qservew4a8kv4quantizationcodesign}
Yujun Lin, Haotian Tang, Shang Yang, Zhekai Zhang, Guangxuan Xiao, Chuang Gan,
  and Song Han.
\newblock Qserve: W4a8kv4 quantization and system co-design for efficient llm
  serving, 2024.

\bibitem{deepseekai2025deepseekr1incentivizingreasoningcapability}
DeepSeek-AI, Daya Guo, Dejian Yang, Haowei Zhang, Junxiao Song, Ruoyu Zhang,
  Runxin Xu, Qihao Zhu, Shirong Ma, Peiyi Wang, Xiao Bi, Xiaokang Zhang,
  Xingkai Yu, Yu~Wu, Z.~F. Wu, Zhibin Gou, Zhihong Shao, Zhuoshu Li, Ziyi Gao,
  Aixin Liu, Bing Xue, Bingxuan Wang, Bochao Wu, Bei Feng, Chengda Lu,
  Chenggang Zhao, Chengqi Deng, Chenyu Zhang, Chong Ruan, Damai Dai, Deli Chen,
  Dongjie Ji, Erhang Li, Fangyun Lin, Fucong Dai, Fuli Luo, Guangbo Hao,
  Guanting Chen, Guowei Li, H.~Zhang, Han Bao, Hanwei Xu, Haocheng Wang,
  Honghui Ding, Huajian Xin, Huazuo Gao, Hui Qu, Hui Li, Jianzhong Guo, Jiashi
  Li, Jiawei Wang, Jingchang Chen, Jingyang Yuan, Junjie Qiu, Junlong Li, J.~L.
  Cai, Jiaqi Ni, Jian Liang, Jin Chen, Kai Dong, Kai Hu, Kaige Gao, Kang Guan,
  Kexin Huang, Kuai Yu, Lean Wang, Lecong Zhang, Liang Zhao, Litong Wang, Liyue
  Zhang, Lei Xu, Leyi Xia, Mingchuan Zhang, Minghua Zhang, Minghui Tang, Meng
  Li, Miaojun Wang, Mingming Li, Ning Tian, Panpan Huang, Peng Zhang, Qiancheng
  Wang, Qinyu Chen, Qiushi Du, Ruiqi Ge, Ruisong Zhang, Ruizhe Pan, Runji Wang,
  R.~J. Chen, R.~L. Jin, Ruyi Chen, Shanghao Lu, Shangyan Zhou, Shanhuang Chen,
  Shengfeng Ye, Shiyu Wang, Shuiping Yu, Shunfeng Zhou, Shuting Pan, S.~S. Li,
  Shuang Zhou, Shaoqing Wu, Shengfeng Ye, Tao Yun, Tian Pei, Tianyu Sun,
  T.~Wang, Wangding Zeng, Wanjia Zhao, Wen Liu, Wenfeng Liang, Wenjun Gao,
  Wenqin Yu, Wentao Zhang, W.~L. Xiao, Wei An, Xiaodong Liu, Xiaohan Wang,
  Xiaokang Chen, Xiaotao Nie, Xin Cheng, Xin Liu, Xin Xie, Xingchao Liu, Xinyu
  Yang, Xinyuan Li, Xuecheng Su, Xuheng Lin, X.~Q. Li, Xiangyue Jin, Xiaojin
  Shen, Xiaosha Chen, Xiaowen Sun, Xiaoxiang Wang, Xinnan Song, Xinyi Zhou,
  Xianzu Wang, Xinxia Shan, Y.~K. Li, Y.~Q. Wang, Y.~X. Wei, Yang Zhang,
  Yanhong Xu, Yao Li, Yao Zhao, Yaofeng Sun, Yaohui Wang, Yi~Yu, Yichao Zhang,
  Yifan Shi, Yiliang Xiong, Ying He, Yishi Piao, Yisong Wang, Yixuan Tan,
  Yiyang Ma, Yiyuan Liu, Yongqiang Guo, Yuan Ou, Yuduan Wang, Yue Gong, Yuheng
  Zou, Yujia He, Yunfan Xiong, Yuxiang Luo, Yuxiang You, Yuxuan Liu, Yuyang
  Zhou, Y.~X. Zhu, Yanhong Xu, Yanping Huang, Yaohui Li, Yi~Zheng, Yuchen Zhu,
  Yunxian Ma, Ying Tang, Yukun Zha, Yuting Yan, Z.~Z. Ren, Zehui Ren, Zhangli
  Sha, Zhe Fu, Zhean Xu, Zhenda Xie, Zhengyan Zhang, Zhewen Hao, Zhicheng Ma,
  Zhigang Yan, Zhiyu Wu, Zihui Gu, Zijia Zhu, Zijun Liu, Zilin Li, Ziwei Xie,
  Ziyang Song, Zizheng Pan, Zhen Huang, Zhipeng Xu, Zhongyu Zhang, and Zhen
  Zhang.
\newblock Deepseek-r1: Incentivizing reasoning capability in llms via
  reinforcement learning, 2025.

\bibitem{qwq32b}
Qwen Team.
\newblock Qwq-32b: Embracing the power of reinforcement learning, March 2025.

\bibitem{DBLP:journals/corr/abs-2412-16720}
Aaron Jaech, Adam Kalai, Adam Lerer, Adam Richardson, Ahmed El{-}Kishky, Aiden
  Low, Alec Helyar, Aleksander Madry, Alex Beutel, Alex Carney, Alex Iftimie,
  Alex Karpenko, Alex~Tachard Passos, Alexander Neitz, Alexander Prokofiev,
  Alexander Wei, Allison Tam, Ally Bennett, Ananya Kumar, Andre Saraiva, Andrea
  Vallone, Andrew Duberstein, Andrew Kondrich, Andrey Mishchenko, Andy
  Applebaum, Angela Jiang, Ashvin Nair, Barret Zoph, Behrooz Ghorbani, Ben
  Rossen, Benjamin Sokolowsky, Boaz Barak, Bob McGrew, Borys Minaiev, Botao
  Hao, Bowen Baker, Brandon Houghton, Brandon McKinzie, Brydon Eastman, Camillo
  Lugaresi, Cary Bassin, Cary Hudson, Chak~Ming Li, Charles de~Bourcy, Chelsea
  Voss, Chen Shen, Chong Zhang, Chris Koch, Chris Orsinger, Christopher Hesse,
  Claudia Fischer, Clive Chan, Dan Roberts, Daniel Kappler, Daniel Levy, Daniel
  Selsam, David Dohan, David Farhi, David Mely, David Robinson, Dimitris
  Tsipras, Doug Li, Dragos Oprica, Eben Freeman, Eddie Zhang, Edmund Wong,
  Elizabeth Proehl, Enoch Cheung, Eric Mitchell, Eric Wallace, Erik Ritter,
  Evan Mays, Fan Wang, Felipe~Petroski Such, Filippo Raso, Florencia Leoni,
  Foivos Tsimpourlas, Francis Song, Fred von Lohmann, Freddie Sulit, Geoff
  Salmon, Giambattista Parascandolo, Gildas Chabot, Grace Zhao, Greg Brockman,
  Guillaume Leclerc, Hadi Salman, Haiming Bao, Hao Sheng, Hart Andrin, Hessam
  Bagherinezhad, Hongyu Ren, Hunter Lightman, Hyung~Won Chung, Ian Kivlichan,
  Ian O'Connell, Ian Osband, Ignasi~Clavera Gilaberte, and Ilge Akkaya.
\newblock Openai o1 system card.
\newblock {\em CoRR}, abs/2412.16720, 2024.

\bibitem{DBLP:journals/corr/SchulmanWDRK17}
John Schulman, Filip Wolski, Prafulla Dhariwal, Alec Radford, and Oleg Klimov.
\newblock Proximal policy optimization algorithms.
\newblock {\em CoRR}, abs/1707.06347, 2017.

\bibitem{DBLP:journals/corr/abs-2402-03300}
Zhihong Shao, Peiyi Wang, Qihao Zhu, Runxin Xu, Junxiao Song, Mingchuan Zhang,
  Y.~K. Li, Y.~Wu, and Daya Guo.
\newblock Deepseekmath: Pushing the limits of mathematical reasoning in open
  language models.
\newblock {\em CoRR}, abs/2402.03300, 2024.

\bibitem{DBLP:conf/acl/AhmadianCGFKPUH24}
Arash Ahmadian, Chris Cremer, Matthias Gall{\'{e}}, Marzieh Fadaee, Julia
  Kreutzer, Olivier Pietquin, Ahmet {\"{U}}st{\"{u}}n, and Sara Hooker.
\newblock Back to basics: Revisiting reinforce-style optimization for learning
  from human feedback in llms.
\newblock In Lun{-}Wei Ku, Andre Martins, and Vivek Srikumar, editors, {\em
  Proceedings of the 62nd Annual Meeting of the Association for Computational
  Linguistics (Volume 1: Long Papers), {ACL} 2024, Bangkok, Thailand, August
  11-16, 2024}, pages 12248--12267. Association for Computational Linguistics,
  2024.

\bibitem{DBLP:conf/icml/LiXZL00L24}
Ziniu Li, Tian Xu, Yushun Zhang, Zhihang Lin, Yang Yu, Ruoyu Sun, and
  Zhi{-}Quan Luo.
\newblock Remax: {A} simple, effective, and efficient reinforcement learning
  method for aligning large language models.
\newblock In {\em Forty-first International Conference on Machine Learning,
  {ICML} 2024, Vienna, Austria, July 21-27, 2024}. OpenReview.net, 2024.

\bibitem{hu2025reinforcesimpleefficientapproach}
Jian Hu.
\newblock Reinforce++: A simple and efficient approach for aligning large
  language models, 2025.

\bibitem{DBLP:conf/nips/RafailovSMMEF23}
Rafael Rafailov, Archit Sharma, Eric Mitchell, Christopher~D. Manning, Stefano
  Ermon, and Chelsea Finn.
\newblock Direct preference optimization: Your language model is secretly a
  reward model.
\newblock In Alice Oh, Tristan Naumann, Amir Globerson, Kate Saenko, Moritz
  Hardt, and Sergey Levine, editors, {\em Advances in Neural Information
  Processing Systems 36: Annual Conference on Neural Information Processing
  Systems 2023, NeurIPS 2023, New Orleans, LA, USA, December 10 - 16, 2023},
  2023.

\bibitem{DBLP:conf/icml/EthayarajhXMJK24}
Kawin Ethayarajh, Winnie Xu, Niklas Muennighoff, Dan Jurafsky, and Douwe Kiela.
\newblock Model alignment as prospect theoretic optimization.
\newblock In {\em Forty-first International Conference on Machine Learning,
  {ICML} 2024, Vienna, Austria, July 21-27, 2024}. OpenReview.net, 2024.

\bibitem{DBLP:conf/emnlp/HongLT24}
Jiwoo Hong, Noah Lee, and James Thorne.
\newblock {ORPO:} monolithic preference optimization without reference model.
\newblock In Yaser Al{-}Onaizan, Mohit Bansal, and Yun{-}Nung Chen, editors,
  {\em Proceedings of the 2024 Conference on Empirical Methods in Natural
  Language Processing, {EMNLP} 2024, Miami, FL, USA, November 12-16, 2024},
  pages 11170--11189. Association for Computational Linguistics, 2024.

\bibitem{DBLP:conf/nips/0001X024}
Yu~Meng, Mengzhou Xia, and Danqi Chen.
\newblock Simpo: Simple preference optimization with a reference-free reward.
\newblock In Amir Globersons, Lester Mackey, Danielle Belgrave, Angela Fan,
  Ulrich Paquet, Jakub~M. Tomczak, and Cheng Zhang, editors, {\em Advances in
  Neural Information Processing Systems 38: Annual Conference on Neural
  Information Processing Systems 2024, NeurIPS 2024, Vancouver, BC, Canada,
  December 10 - 15, 2024}, 2024.

\bibitem{DBLP:conf/icml/ChenDYJG24}
Zixiang Chen, Yihe Deng, Huizhuo Yuan, Kaixuan Ji, and Quanquan Gu.
\newblock Self-play fine-tuning converts weak language models to strong
  language models.
\newblock In {\em Forty-first International Conference on Machine Learning,
  {ICML} 2024, Vienna, Austria, July 21-27, 2024}. OpenReview.net, 2024.

\bibitem{DBLP:journals/corr/abs-2406-01660}
Eugene Choi, Arash Ahmadian, Matthieu Geist, Olivier Pietquin, and
  Mohammad~Gheshlaghi Azar.
\newblock Self-improving robust preference optimization.
\newblock {\em CoRR}, abs/2406.01660, 2024.

\bibitem{DBLP:journals/corr/abs-2408-06195}
Zhenting Qi, Mingyuan Ma, Jiahang Xu, Li~Lyna Zhang, Fan Yang, and Mao Yang.
\newblock Mutual reasoning makes smaller llms stronger problem-solvers.
\newblock {\em CoRR}, abs/2408.06195, 2024.

\bibitem{chen2019bertjointintentclassification}
Qian Chen, Zhu Zhuo, and Wen Wang.
\newblock Bert for joint intent classification and slot filling, 2019.

\bibitem{DBLP:conf/epia/TavaresASSM23}
Diogo Tavares, Pedro Azevedo, David Semedo, Ricardo~Gamelas Sousa, and
  Jo{\~{a}}o Magalh{\~{a}}es.
\newblock Task conditioned {BERT} for joint intent detection and slot-filling.
\newblock In Nuno Moniz, Zita Vale, Jos{\'{e}} Cascalho, Catarina Silva, and
  Raquel Sebasti{\~{a}}o, editors, {\em Progress in Artificial Intelligence -
  22nd {EPIA} Conference on Artificial Intelligence, {EPIA} 2023, Faial Island,
  Azores, September 5-8, 2023, Proceedings, Part {I}}, volume 14115 of {\em
  Lecture Notes in Computer Science}, pages 467--480. Springer, 2023.

\bibitem{lewis2021retrievalaugmentedgenerationknowledgeintensivenlp}
Patrick Lewis, Ethan Perez, Aleksandra Piktus, Fabio Petroni, Vladimir
  Karpukhin, Naman Goyal, Heinrich Küttler, Mike Lewis, Wen tau Yih, Tim
  Rocktäschel, Sebastian Riedel, and Douwe Kiela.
\newblock Retrieval-augmented generation for knowledge-intensive nlp tasks,
  2021.

\bibitem{li2023towards}
Zehan Li, Xin Zhang, Yanzhao Zhang, Dingkun Long, Pengjun Xie, and Meishan
  Zhang.
\newblock Towards general text embeddings with multi-stage contrastive
  learning.
\newblock {\em arXiv preprint arXiv:2308.03281}, 2023.

\bibitem{johnson2019billion}
Jeff Johnson, Matthijs Douze, and Herv{\'e} J{\'e}gou.
\newblock Billion-scale similarity search with {GPUs}.
\newblock {\em IEEE Transactions on Big Data}, 7(3):535--547, 2019.

\bibitem{HintonSalakhutdinov2006b}
G~E Hinton and R~R Salakhutdinov.
\newblock Reducing the dimensionality of data with neural networks.
\newblock {\em Science}, 313(5786):504--507, July 2006.

\bibitem{sreenivas2024llmpruningdistillationpractice}
Sharath~Turuvekere Sreenivas, Saurav Muralidharan, Raviraj Joshi, Marcin
  Chochowski, Ameya~Sunil Mahabaleshwarkar, Gerald Shen, Jiaqi Zeng, Zijia
  Chen, Yoshi Suhara, Shizhe Diao, Chenhan Yu, Wei-Chun Chen, Hayley Ross,
  Oluwatobi Olabiyi, Ashwath Aithal, Oleksii Kuchaiev, Daniel Korzekwa, Pavlo
  Molchanov, Mostofa Patwary, Mohammad Shoeybi, Jan Kautz, and Bryan Catanzaro.
\newblock Llm pruning and distillation in practice: The minitron approach,
  2024.

\bibitem{meta2024llama32}
Meta.
\newblock {Llama 3.2: Vision and Lightweight Models for Edge and Mobile
  Devices}.
\newblock
  \url{https://ai.meta.com/blog/llama-3-2-connect-2024-vision-edge-mobile-devices/},
  January 2024.
\newblock Accessed: 2025-03-14.

\bibitem{gunter2024appleintelligencefoundationlanguage}
Tom Gunter, Zirui Wang, Chong Wang, Ruoming Pang, Andy Narayanan, Aonan Zhang,
  Bowen Zhang, Chen Chen, Chung-Cheng Chiu, David Qiu, Deepak Gopinath,
  Dian~Ang Yap, Dong Yin, Feng Nan, Floris Weers, Guoli Yin, Haoshuo Huang,
  Jianyu Wang, Jiarui Lu, John Peebles, Ke~Ye, Mark Lee, Nan Du, Qibin Chen,
  Quentin Keunebroek, Sam Wiseman, Syd Evans, Tao Lei, Vivek Rathod, Xiang
  Kong, Xianzhi Du, Yanghao Li, Yongqiang Wang, Yuan Gao, Zaid Ahmed, Zhaoyang
  Xu, Zhiyun Lu, Al~Rashid, Albin~Madappally Jose, Alec Doane, Alfredo Bencomo,
  Allison Vanderby, Andrew Hansen, Ankur Jain, Anupama~Mann Anupama, Areeba
  Kamal, Bugu Wu, Carolina Brum, Charlie Maalouf, Chinguun Erdenebileg, Chris
  Dulhanty, Dominik Moritz, Doug Kang, Eduardo Jimenez, Evan Ladd, Fangping
  Shi, Felix Bai, Frank Chu, Fred Hohman, Hadas Kotek, Hannah~Gillis Coleman,
  Jane Li, Jeffrey Bigham, Jeffery Cao, Jeff Lai, Jessica Cheung, Jiulong Shan,
  Joe Zhou, John Li, Jun Qin, Karanjeet Singh, Karla Vega, Kelvin Zou, Laura
  Heckman, Lauren Gardiner, Margit Bowler, Maria Cordell, Meng Cao, Nicole Hay,
  Nilesh Shahdadpuri, Otto Godwin, Pranay Dighe, Pushyami Rachapudi, Ramsey
  Tantawi, Roman Frigg, Sam Davarnia, Sanskruti Shah, Saptarshi Guha, Sasha
  Sirovica, Shen Ma, Shuang Ma, Simon Wang, Sulgi Kim, Suma Jayaram, Vaishaal
  Shankar, Varsha Paidi, Vivek Kumar, Xin Wang, Xin Zheng, Walker Cheng, Yael
  Shrager, Yang Ye, Yasu Tanaka, Yihao Guo, Yunsong Meng, Zhao~Tang Luo, Zhi
  Ouyang, Alp Aygar, Alvin Wan, Andrew Walkingshaw, Andy Narayanan, Antonie
  Lin, Arsalan Farooq, Brent Ramerth, Colorado Reed, Chris Bartels, Chris
  Chaney, David Riazati, Eric~Liang Yang, Erin Feldman, Gabriel Hochstrasser,
  Guillaume Seguin, Irina Belousova, Joris Pelemans, Karen Yang,
  Keivan~Alizadeh Vahid, Liangliang Cao, Mahyar Najibi, Marco Zuliani, Max
  Horton, Minsik Cho, Nikhil Bhendawade, Patrick Dong, Piotr Maj, Pulkit
  Agrawal, Qi~Shan, Qichen Fu, Regan Poston, Sam Xu, Shuangning Liu, Sushma
  Rao, Tashweena Heeramun, Thomas Merth, Uday Rayala, Victor Cui,
  Vivek~Rangarajan Sridhar, Wencong Zhang, Wenqi Zhang, Wentao Wu, Xingyu Zhou,
  Xinwen Liu, Yang Zhao, Yin Xia, Zhile Ren, and Zhongzheng Ren.
\newblock Apple intelligence foundation language models, 2024.

\bibitem{tensorrt-llm}
NVIDIA.
\newblock Tensorrt-llm.
\newblock \url{https://github.com/NVIDIA/TensorRT-LLM}, 2023.

\bibitem{DBLP:conf/naacl/DevlinCLT19}
Jacob Devlin, Ming{-}Wei Chang, Kenton Lee, and Kristina Toutanova.
\newblock {BERT:} pre-training of deep bidirectional transformers for language
  understanding.
\newblock In Jill Burstein, Christy Doran, and Thamar Solorio, editors, {\em
  Proceedings of the 2019 Conference of the North American Chapter of the
  Association for Computational Linguistics: Human Language Technologies,
  {NAACL-HLT} 2019, Minneapolis, MN, USA, June 2-7, 2019, Volume 1 (Long and
  Short Papers)}, pages 4171--4186. Association for Computational Linguistics,
  2019.

\bibitem{DBLP:conf/nips/LiuHZZLKTYLFNYS24}
Zuxin Liu, Thai Hoang, Jianguo Zhang, Ming Zhu, Tian Lan, Shirley Kokane,
  Juntao Tan, Weiran Yao, Zhiwei Liu, Yihao Feng, Rithesh~R. N., Liangwei Yang,
  Silvio Savarese, Juan~Carlos Niebles, Huan Wang, Shelby Heinecke, and Caiming
  Xiong.
\newblock Apigen: Automated pipeline for generating verifiable and diverse
  function-calling datasets.
\newblock In Amir Globersons, Lester Mackey, Danielle Belgrave, Angela Fan,
  Ulrich Paquet, Jakub~M. Tomczak, and Cheng Zhang, editors, {\em Advances in
  Neural Information Processing Systems 38: Annual Conference on Neural
  Information Processing Systems 2024, NeurIPS 2024, Vancouver, BC, Canada,
  December 10 - 15, 2024}, 2024.

\bibitem{berkeley-function-calling-leaderboard}
Fanjia Yan, Huanzhi Mao, Charlie Cheng-Jie Ji, Tianjun Zhang, Shishir~G. Patil,
  Ion Stoica, and Joseph~E. Gonzalez.
\newblock Berkeley function calling leaderboard.
\newblock
  \url{https://gorilla.cs.berkeley.edu/blogs/8_berkeley_function_calling_leaderboard.html},
  2024.

\bibitem{yang2024qwen2technicalreport}
An~Yang, Baosong Yang, Binyuan Hui, Bo~Zheng, Bowen Yu, Chang Zhou, Chengpeng
  Li, Chengyuan Li, Dayiheng Liu, Fei Huang, Guanting Dong, Haoran Wei, Huan
  Lin, Jialong Tang, Jialin Wang, Jian Yang, Jianhong Tu, Jianwei Zhang,
  Jianxin Ma, Jianxin Yang, Jin Xu, Jingren Zhou, Jinze Bai, Jinzheng He,
  Junyang Lin, Kai Dang, Keming Lu, Keqin Chen, Kexin Yang, Mei Li, Mingfeng
  Xue, Na~Ni, Pei Zhang, Peng Wang, Ru~Peng, Rui Men, Ruize Gao, Runji Lin,
  Shijie Wang, Shuai Bai, Sinan Tan, Tianhang Zhu, Tianhao Li, Tianyu Liu,
  Wenbin Ge, Xiaodong Deng, Xiaohuan Zhou, Xingzhang Ren, Xinyu Zhang, Xipin
  Wei, Xuancheng Ren, Xuejing Liu, Yang Fan, Yang Yao, Yichang Zhang, Yu~Wan,
  Yunfei Chu, Yuqiong Liu, Zeyu Cui, Zhenru Zhang, Zhifang Guo, and Zhihao Fan.
\newblock Qwen2 technical report, 2024.

\bibitem{DBLP:conf/iclr/LoshchilovH19}
Ilya Loshchilov and Frank Hutter.
\newblock Decoupled weight decay regularization.
\newblock In {\em 7th International Conference on Learning Representations,
  {ICLR} 2019, New Orleans, LA, USA, May 6-9, 2019}. OpenReview.net, 2019.

\bibitem{Liu2025KL}
YiMing Liu.
\newblock Rethinking kl divergence in rlhf: From single sample to mini-batch to
  expectation, 2025.
\newblock Notion Blog.

\bibitem{DBLP:conf/sc/NarayananSCLPKV21}
Deepak Narayanan, Mohammad Shoeybi, Jared Casper, Patrick LeGresley, Mostofa
  Patwary, Vijay Korthikanti, Dmitri Vainbrand, Prethvi Kashinkunti, Julie
  Bernauer, Bryan Catanzaro, Amar Phanishayee, and Matei Zaharia.
\newblock Efficient large-scale language model training on {GPU} clusters using
  megatron-lm.
\newblock In Bronis~R. de~Supinski, Mary~W. Hall, and Todd Gamblin, editors,
  {\em International Conference for High Performance Computing, Networking,
  Storage and Analysis, {SC} 2021, St. Louis, Missouri, USA, November 14-19,
  2021}, page~58. {ACM}, 2021.

\bibitem{DBLP:journals/corr/abs-2405-11143}
Jian Hu, Xibin Wu, Weixun Wang, Xianyu, Dehao Zhang, and Yu~Cao.
\newblock Openrlhf: An easy-to-use, scalable and high-performance {RLHF}
  framework.
\newblock {\em CoRR}, abs/2405.11143, 2024.

\bibitem{kwon2023efficient}
Woosuk Kwon, Zhuohan Li, Siyuan Zhuang, Ying Sheng, Lianmin Zheng, Cody~Hao Yu,
  Joseph~E. Gonzalez, Hao Zhang, and Ion Stoica.
\newblock Efficient memory management for large language model serving with
  pagedattention.
\newblock In {\em Proceedings of the ACM SIGOPS 29th Symposium on Operating
  Systems Principles}, 2023.

\bibitem{zheng2024sglangefficientexecutionstructured}
Lianmin Zheng, Liangsheng Yin, Zhiqiang Xie, Chuyue Sun, Jeff Huang, Cody~Hao
  Yu, Shiyi Cao, Christos Kozyrakis, Ion Stoica, Joseph~E. Gonzalez, Clark
  Barrett, and Ying Sheng.
\newblock Sglang: Efficient execution of structured language model programs,
  2024.

\bibitem{InfinityInstruct2024}
Beijing~Academy of~Artificial Intelligence~(BAAI).
\newblock Infinity instruct.
\newblock {\em arXiv preprint arXiv:2406.XXXX}, 2024.

\bibitem{hendrycks2021measuringmassivemultitasklanguage}
Dan Hendrycks, Collin Burns, Steven Basart, Andy Zou, Mantas Mazeika, Dawn
  Song, and Jacob Steinhardt.
\newblock Measuring massive multitask language understanding, 2021.

\bibitem{li2024cmmlumeasuringmassivemultitask}
Haonan Li, Yixuan Zhang, Fajri Koto, Yifei Yang, Hai Zhao, Yeyun Gong, Nan
  Duan, and Timothy Baldwin.
\newblock Cmmlu: Measuring massive multitask language understanding in chinese,
  2024.

\bibitem{clark2018thinksolvedquestionanswering}
Peter Clark, Isaac Cowhey, Oren Etzioni, Tushar Khot, Ashish Sabharwal, Carissa
  Schoenick, and Oyvind Tafjord.
\newblock Think you have solved question answering? try arc, the ai2 reasoning
  challenge, 2018.

\bibitem{zellers2019hellaswagmachinereallyfinish}
Rowan Zellers, Ari Holtzman, Yonatan Bisk, Ali Farhadi, and Yejin Choi.
\newblock Hellaswag: Can a machine really finish your sentence?, 2019.

\bibitem{cobbe2021trainingverifierssolvemath}
Karl Cobbe, Vineet Kosaraju, Mohammad Bavarian, Mark Chen, Heewoo Jun, Lukasz
  Kaiser, Matthias Plappert, Jerry Tworek, Jacob Hilton, Reiichiro Nakano,
  Christopher Hesse, and John Schulman.
\newblock Training verifiers to solve math word problems, 2021.

\bibitem{korthikanti2022reducingactivationrecomputationlarge}
Vijay Korthikanti, Jared Casper, Sangkug Lym, Lawrence McAfee, Michael
  Andersch, Mohammad Shoeybi, and Bryan Catanzaro.
\newblock Reducing activation recomputation in large transformer models, 2022.

\bibitem{lin2025robust}
Qiqiang Lin, Muning Wen, Qiuying Peng, Guanyu Nie, Junwei Liao, Jun Wang,
  Xiaoyun Mo, Jiamu Zhou, Cheng Cheng, Yin Zhao, Jun Wang, and Weinan Zhang.
\newblock Robust function-calling for on-device language model via function
  masking.
\newblock In {\em The Thirteenth International Conference on Learning
  Representations}, 2025.

\bibitem{colvin2025pydantic}
Samuel Colvin, Eric Jolibois, Hasan Ramezani, Adrian Garcia~Badaracco, Terrence
  Dorsey, David Montague, Serge Matveenko, Marcelo Trylesinski, Sydney Runkle,
  David Hewitt, Alex Hall, and Victorien Plot.
\newblock Pydantic.
\newblock \url{https://github.com/pydantic/pydantic}, 2025.
\newblock License: MIT.

\end{thebibliography}
\end{document}